\documentclass{bmvc2k}

\title{Casual Indoor HDR Radiance Capture from Omnidirectional Images}

\addauthor{Pulkit Gera}{pulkit.gera@research.iiit.ac.in}{1}
\addauthor{Mohammad Reza Karimi Dastjerdi}{mohammad.karimi-dastjerdi.1@ulaval.ca}{2}
\addauthor{Charles Renaud}{charles.renaud.1@ulaval.ca}{2}
\addauthor{P. J. Narayanan}{pjn@iiit.ac.in}{1}
\addauthor{Jean-François Lalonde}{jflalonde@gel.ulaval.ca}{2}

\addinstitution{
 Centre for Visual Information Technology \& 
 Kohli Centre on Intelligent Systems (KCIS)\\
 International Institute of Information Technology (IIIT),\\
 Hyderabad, India
}
\addinstitution{
 Université Laval\\
 Quebec City, Canada
}

\runninghead{Gera, Karimi, Renaud, PJN, Lalonde}{PanoHDR-NeRF}
\usepackage{microtype}
\usepackage{booktabs}
\usepackage{multirow}
\usepackage{siunitx}
\usepackage{xspace}
\usepackage{nicefrac}
\usepackage[export]{adjustbox}
\usepackage{cleveref}
\crefname{section}{sec.}{secs.}
\Crefname{section}{Sec.}{Secs.}
\crefname{table}{tab.}{tabs.}
\Crefname{table}{Tab.}{Tabs.}
\crefname{figure}{fig.}{figs.}
\Crefname{figure}{Fig.}{Figs.}
\crefname{equation}{eq.}{eqs.}
\Crefname{equation}{Eq.}{Eqs.}
\newcommand{\etal}{\emph{et al.\xspace}}

\usepackage{xcolor}

\newcommand{\methodname}{PanoHDR-NeRF\xspace}

\def\etal{\emph{et al}\bmvaOneDot}

\begin{document}

\maketitle
\begin{abstract}
We present \methodname, a neural representation of the full HDR radiance field of an indoor scene, and a pipeline to capture it casually, without elaborate setups or complex capture protocols. First, a user captures a low dynamic range (LDR) omnidirectional video of the scene by freely waving an off-the-shelf camera around the scene. Then, an LDR2HDR network uplifts the captured LDR frames to HDR, which are used to train a tailored NeRF++ model. The resulting \methodname can render full HDR images from any location of the scene. Through experiments on a novel test dataset of real scenes with the ground truth HDR radiance captured at locations not seen during training, we show that \methodname predicts plausible HDR radiance from any scene point. We also show that the predicted radiance can synthesize correct lighting effects, enabling the augmentation of indoor scenes with synthetic objects that are lit correctly. Datasets and code are available at {\footnotesize \url{https://lvsn.github.io/PanoHDR-NeRF/}}.

\end{abstract}



\section{Introduction}

\begin{figure} [t]
\centering
  \includegraphics[width=0.9\textwidth]{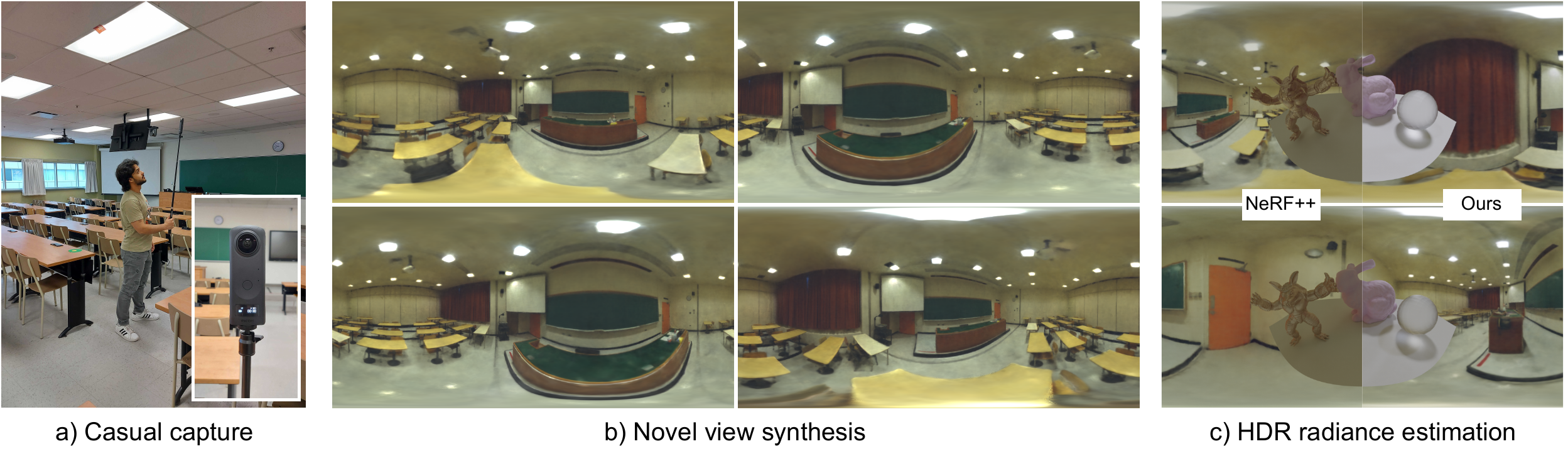}
  \caption{We capture the continuous HDR radiance of an indoor scene. Our \methodname approach takes a) casually captured LDR images from an off-the-shelf camera (shown in inset) as input, and performs b) novel view synthesis of the indoor scene. c) As opposed to existing techniques such as NeRF++~\cite{nerf++} (left), \methodname (right) properly estimates the HDR radiance of the scene, visualized by relighting virtual test objects.}
  \label{fig:teaser}
\end{figure}

Capturing the incoming radiance of a given scene is an important step for many augmented reality applications, since it allows immersive exploration and realistic scene augmentation. To accurately measure the radiance, Debevec~\cite{debevec-sig-98} pioneered \emph{image-based lighting} which involves photographing a chrome sphere---called a {\em light probe}---at different exposures and merging them into a single, high dynamic range (HDR) image~\cite{debevec-siggraph-97}. HDR light probing was later extended to use wide angle lenses or $360^\circ$ cameras and is at the heart of the lighting capture necessary to achieve special effects in movies today\footnote{\scriptsize \url{https://www.fxguide.com/fxfeatured/the-definitive-weta-digital-guide-to-ibl/}}. Because of the close proximity between light sources and objects in the scene, the HDR radiance field of a typical indoor scene varies rapidly: lighting near a window is vastly different from the center of the room. Accurately capturing indoor radiance involves moving the apparatus and repeating the operation many times, limiting scalability. Approximations such as reprojecting the measured radiance onto a proxy 3D model~\cite{debevec1998efficient} must be used. To simplify the capture process, one could use inverse tonemapping techniques~\cite{rempel-siggraph-07,reversetone,brown2019} which estimate the true dynamic range information from low dynamic range (LDR) inputs. This yields radiance estimates only where images are taken. Can we capture and learn a representation of the scene from which HDR radiance can be synthesized at any point?

Novel view synthesis of an object or scene from multiple views has garnered attention recently, especially since the emergence of neural radiance fields (NeRF)~\cite{nerf}. In this approach, a set of images are captured, co-registered using structure-from-motion (SfM), and used to train a deep network that learns to predict the color and opacity along any 3D ray using volumetric rendering. Recent flavors of NeRF model large scenes~\cite{nerf++,barron2022mipnerf360}, handle omnidirectional input images~\cite{omni-fish}, and decompose the scene into intrinsic components~\cite{nerd,nerfactor}. Most NeRF-based methods accept LDR images as input and do not accurately model the true HDR radiance of indoor environments. There are recent attempts to address this limitation~\cite{hdrnerf,rawnerf}, which require multiple LDR images at different exposures to model the full dynamic range of indoor lighting. Capturing 11 exposures at a point of the scene, which is necessary to reconstruct a reasonable HDR panorama spanning over 22 f-stops, takes approximately two minutes with a conventional $360^\circ$ camera (Ricoh Theta Z1). Thus to capture HDR lighting for a large indoor scene, we need over 200 images bringing the capture time to over 3.5 hours. In addition, an elaborate setup would be required to avoid ghosting artifacts. These methods to capture the full dynamic range of lighting for large indoor scenes require specialized setup and are tedious and cumbersome.


In this paper, we present \methodname, a neural representation of the plausible full HDR radiance field of an indoor scene. We also present a method to generate it from casually captured images of the scene. HDR radiance from any novel viewpoint in the scene can subsequently be estimated (\cref{fig:teaser}) from \methodname. Our method does not require any special equipment or complicated capture protocols. It accepts as input a video sequence captured by freely moving a commercial $360^\circ$ camera around the scene. As output, it can produce the HDR radiance at any given location in the scene. To do so, we leverage two deep neural networks: 1) an LDR2HDR model that predicts the HDR radiance from a single LDR panorama captured by the camera and 2) a modified NeRF++ model trained on the predicted HDR outputs of the first network. To evaluate our proposed method, we capture a set of six different indoor scenes, which we augment with a set of ground truth HDR light probes at each scene. Our experiments demonstrate that, despite the simplicity of the capture procedure, \methodname can accurately predict HDR radiance in a variety of challenging conditions. Our approach can render $360^\circ$ spatially varying HDR light probes, which can be used to provide correct lighting effects when the scene is augmented with virtual objects. Compared to previous methods, our approach reduces the complexity and time required to capture the full HDR radiance field of the scene.

\section{Related work}

\noindent\textbf{Inverse tonemapping} \quad Inverse tonemapping aims to recover missing information in the over- and under-saturated areas of an LDR image. While earlier methods~\cite{rempel-siggraph-07} relied on heuristics, several deep learning architectures have been proposed recently~\cite{reversetone,rendering}. These include a 2D encoder and 3D decoder with skip connections to generate bracketed LDR image stacks (over exposed and under exposed)~\cite{reversetone}, an encoder-decoder to reconstruct the HDR image directly~\cite{hdrnet}, and a multiscale autoencoder architecture to learn multilevel features from LDR image which are merged to reconstruct HDR images~\cite{expandnet}. Lee \etal~\cite{deepchain} generate LDR stacks with a two-branch network. DeepHDR~\cite{deephdr} mask out the saturated areas to reduce sub-optimal features from well-exposed and saturated pixels.
HallucinationNet~\cite{hallu} individually models the main components of the imaging pipeline: dynamic range clipping, camera response function, and quantization. 
LANet~\cite{lanet} introduces a multi-task network with a luminance attention and HDR reconstruction streams. 
We build on the LANet architecture and augment it with an additional rendering loss. This combination outperforms alternatives at predicting the high dynamic range radiance in indoor environments. 

\noindent\textbf{Novel view synthesis} \quad Classical methods reconstruct an explicit 3D model of the scene~\cite{debevec1998efficient,GortlerGSC96,HedmanRDB16}. Recent works utilize off-the-shelf SfM techniques to generate a coarse geometry and use neural networks to render photorealistic novel views \cite{hedman18siggraph,meshry19cvpr,dnr}. Breaking the scene into Multi-Plane Image (MPI) representation to render novel views by blending has been tried~\cite{flynn19cvpr,stereo18tog,llff}. DeepVoxels~\cite{sitzmann19cvpr} learn a voxel-based volumetric representation of the scene using Gated Recurrent Units (GRUs)~\cite{DBLP:conf/emnlp/ChoMGBBSB14}. Free View Synthesis and Stable View Synthesis ~\cite{Riegler2020FVS,Riegler2021StableVS} map the encoded features from the source images into the target view and blend them via a neural network directly or in a geometric space.
Novel view synthesis from omnidirectional images are fairly limited so far. Huang \etal~\cite{huangpano} reconstruct a point cloud from $360^\circ$ videos to achieve real time video playback on a VR device. Serrano \etal~\cite{serrano} design a layered scene representation that facilitates parallax and real time playback of $360^\circ$ video. \cite{linpano,attalpano} present Multi-Depth Panorama (MDP) and multi-sphere images to create 6-DoF renderings. This requires an elaborate setup and does not accommodate free viewpoint synthesis. Bertel \etal~\cite{omni_photo} propose a fast, casual and robust capture of immersive real-world VR experience. However it takes a lot of memory and poor proxy geometry causes warping artifacts. Xu \etal~\cite{xupano} estimate the entire indoor scene from a single image using a CNN but need room layout priors and depth that are challenging to obtain for real scenes. 

\noindent\textbf{Neural radiance fields} \quad Neural Radiance Fields (NeRF)~\cite{nerf} learn an implicit volumetric scene representation with a Multi-Layer Perceptron (MLP) that receives viewing positions and directions as input and predicts the RGB colour and opacity as output. The original formulation was later extended in several ways relevant to our work.
NeRF++~\cite{nerf++}, which we leverage, models an unbounded scene by splitting it into foreground/background, learning each separately. Mega-NeRF~\cite{meganerf} divides the scene into smaller sections. Block-NeRF~\cite{blocknerf} splits city-scale scenes into blocks, trains a NeRF for each block, and combines the results. Mip-NeRF~\cite{mipnerf,barron2022mipnerf360} replaces rays with anti-aliased conical frustums for speed and accuracy.
NeRF with HDR images has been explored in two works. NeRF in the dark~\cite{rawnerf} train directly on linear RAW images with higher effective dynamic range (14-16 bits compared to more typical 8-bit cameras). 
Huang \etal~\cite{hdrnerf} train from a set of LDR images captured at alternating exposures. Both methods require multiple LDR images at different exposures, which is time-consuming to capture and requires specialized setup. In contrast, our method plausibly predicts the full dynamic range from a single casually captured video.
%
Finally, panoramic images have been explored in \cite{omninerf,omni++}, who present a method to synthesize panoramas from a single omnidirectional input. They require depth as input and novel views can only be rendered on a straight line path. CylindricalNeRF~\cite{cylinder_nerf} proposes cylindrical sampling for unbounded scenes captured in circular trajectory. OmniNeRF~\cite{omni-fish} synthesizes novel fish-eye projection images, using spherical sampling to improve the quality of results as we do.

\begin{figure*}[t]
    \centering
    \includegraphics[width=\linewidth]{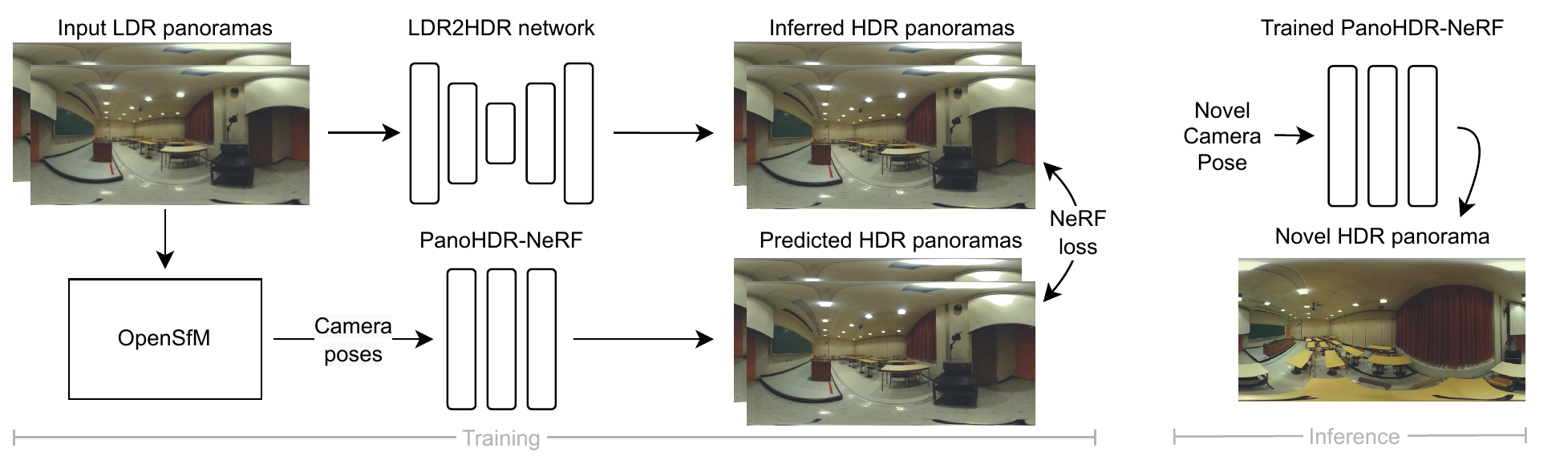}
    \caption{Overview of our pipeline. At training time (left), the captured panoramas are linearized (calibrated using a color checker, not shown) and processed by a pre-trained LDR2HDR network to obtain HDR estimates. The HDR panoramas, along with the camera poses obtained with OpenSfM~\cite{opensfm}, are used to train the \methodname network, which learns to synthesize HDR scene radiance at any point in the scene. At inference time (right), we simply provide the novel camera pose and obtain the corresponding novel HDR panorama.}
    \vspace*{-5mm}
    \label{fig:overview}
\end{figure*}
%

\noindent\textbf{HDR scene capture} \quad Zhang \etal~\cite{zhang2016emptying} leverage an RGBD scan of an indoor scene and estimate scene parameters including HDR radiance over scene geometry. Walton \etal~\cite{waltondynamic} combine a depth sensor with a fisheye camera in a SLAM-based approach to recover geometry and lighting. Tarko \etal~\cite{brown2019} takes an omnidirectional video as input and uses inverse tone mapping~\cite{reversetone} to convert it to HDR. Yang \etal~\cite{yangneural} learn background and objects as separate NeRF models and combine them with a captured panoramic image into a single scene. Our method synthesizes panoramas from novel viewpoints with full dynamic range. 




\section{Method}

Given a set of LDR panoramas $\{I_k\}_{k=1}^{N}$ captured freely using an off-the-shelf 360$^\circ$ camera and associated camera poses $\{\mathbf{P}_k\}_{k=1}^{N}$ obtained with SfM, our objective is to predict the HDR radiance of the scene at any novel viewpoint. We achieve this by recovering HDR values from LDR frames using learning-based inverse tone mapping and then using them as the supervision for novel view synthesis. An overview of our method is given in \cref{fig:overview}. 

\subsection{High dynamic range with LDR2HDR network}
\label{sec:ldr2hdr}

Extrapolating HDR from LDR inputs is typically framed as recovering values in the over- and under-exposed regions. Our work focuses on recovering the over-exposed regions exclusively, with the goal of predicting accurate HDR radiance values (specially the intensities of light sources) for realistic virtual object insertion.

\noindent\textbf{Network architecture} \quad We borrow the network architecture proposed in Luminance Attentive Networks (LANet)~\cite{lanet}, which is designed as a multi-task network with two \emph{luminance attention} and \emph{HDR reconstruction} streams. The former attempts to create a spatially-weighted attention map of the over-exposed regions in the input image, while the latter uses the attention map to estimate the HDR images. Note that we do not use their proposed adaptation for panoramas since it did not improve the performance in our case. So, we train the model on equirectangular images directly. 

\noindent\textbf{Loss functions} \quad For the LDR2HDR module, we use the same loss function $\ell_\mathrm{lanet}$ as \cite{lanet}, combining scale invariant and luminance segmentation losses. The HDR panoramas are radiance values used to light the scene. To match rendering quality, we use a rendering loss
\begin{equation}
\ell_\mathrm{rend} = || \mathbf{T}\ \mathbf{y}_\mathrm{HDR} - \mathbf{T}\ \mathbf{t}_\mathrm{HDR} ||_2 \,,
\label{eqn:loss-render}
\end{equation}	
where $\mathbf{T}$ is a pre-computed transport \cite{prt} matrix for a Lambertian scene (single bounce), $\mathbf{y}_\mathrm{HDR}$ is the predicted HDR panorama, and $\mathbf{t}_\mathrm{HDR}$ is the ground truth. A top-down view of a ``spiky sphere`` on a plane is used for $\mathbf{T}$. The rendered images guide the network about the color and direction of HDR lighting in the scene.
The final loss for training the LDR2HDR network is an equally weighted combination $\ell_\mathrm{hdr} = \ell_\mathrm{lanet} + \ell_\mathrm{rend}$. \\

\noindent\textbf{Datasets} \quad We pretrain the LDR2HDR network on the Laval Indoor HDR Dataset~\cite{Gardner2017LearningTP}, which consists of 2,400 HDR panoramas captured in a variety of indoor settings, with a train, validation and test split as 80:10:10. Since the sensor used to capture it (Canon 5D Mark iii camera) and the sensor we use to capture indoor scenes (Ricoh Theta Z1) are different, a domain gap was observed. To alleviate this, we finetune the LDR2HDR network on a small dataset of 78 HDR panoramas captured at different locations, different from the test scenes, using the target sensor (Theta Z1).

\subsection{Continuous HDR radiance with PanoHDR-NeRF}
\label{sec:panohdr-nerf}


\noindent\textbf{Network architecture} \quad We take inspiration from and combine several recent works on NeRF. First, we employ the NeRF++ architecture as the base. Second, similar to \cite{barron2022mipnerf360}, we incorporate the anti-aliased conical frustums from Mip-NeRF~\cite{mipnerf}. We train \methodname by sampling rays in spherical coordinates instead of pixel coordinates (more details in supplementary).

\begin{figure}
\centering
\setlength{\tabcolsep}{1pt}
\footnotesize
\begin{tabular}{cccc}
Cafeteria & Chess room & Dark class & Spotlights \\
\includegraphics[width=0.248\linewidth]{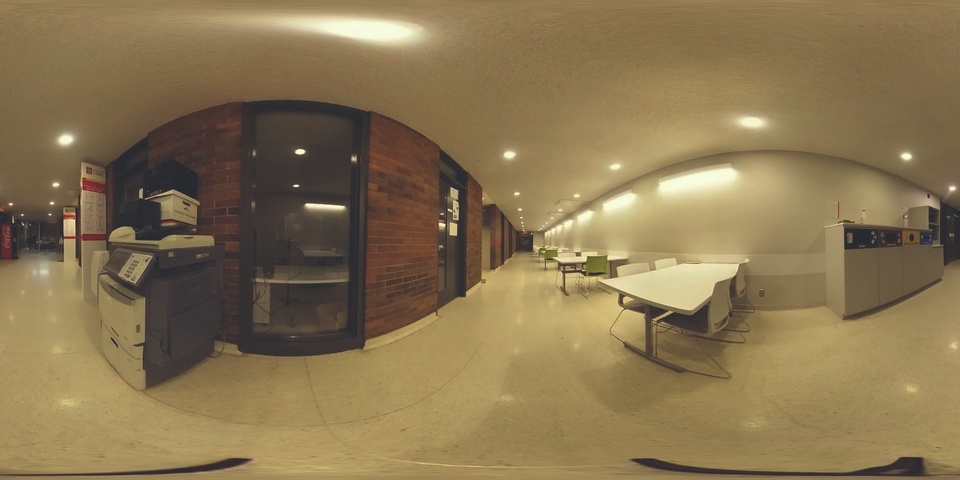} &
\includegraphics[width=0.248\linewidth]{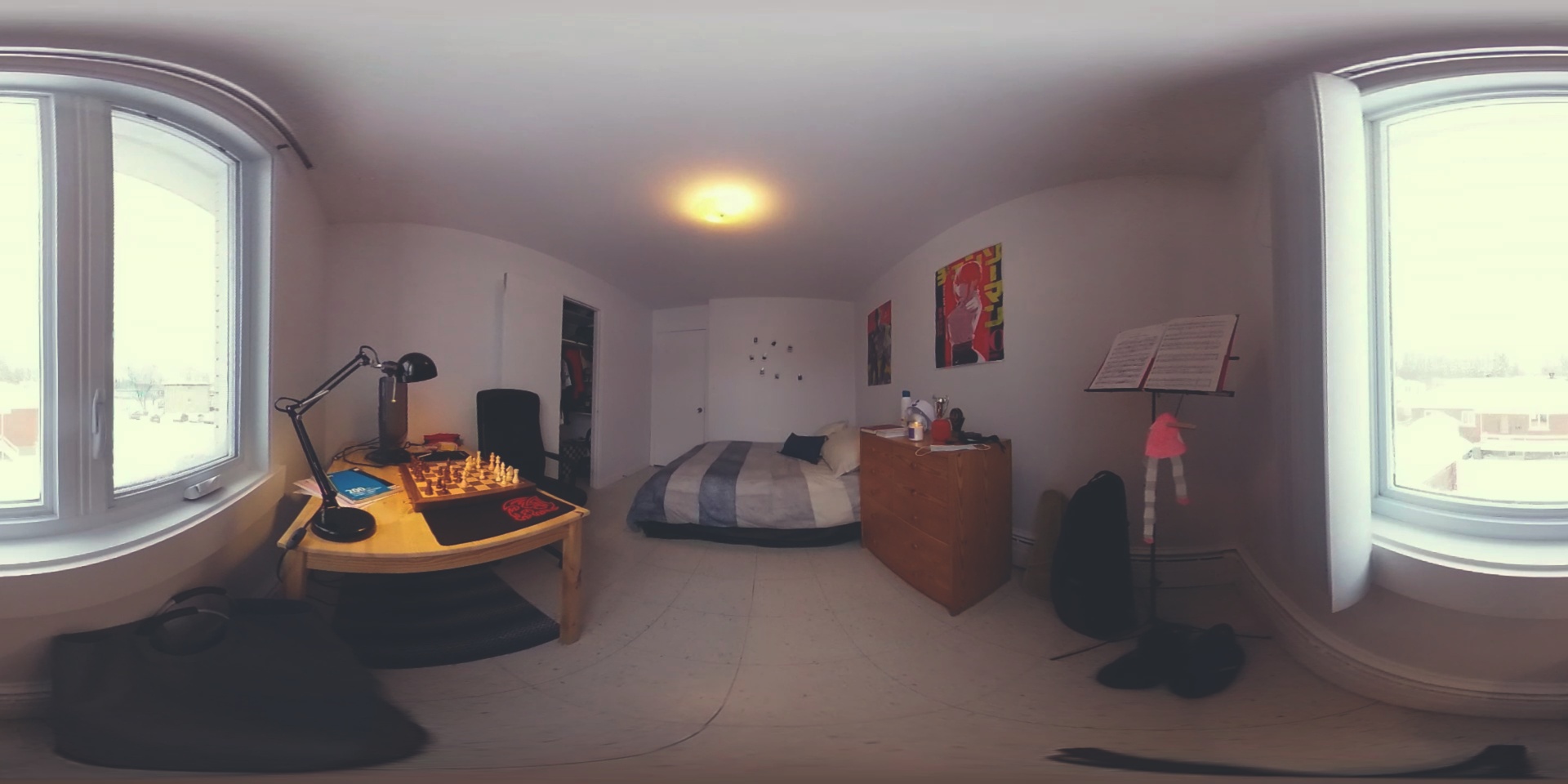} &
\includegraphics[width=0.248\linewidth]{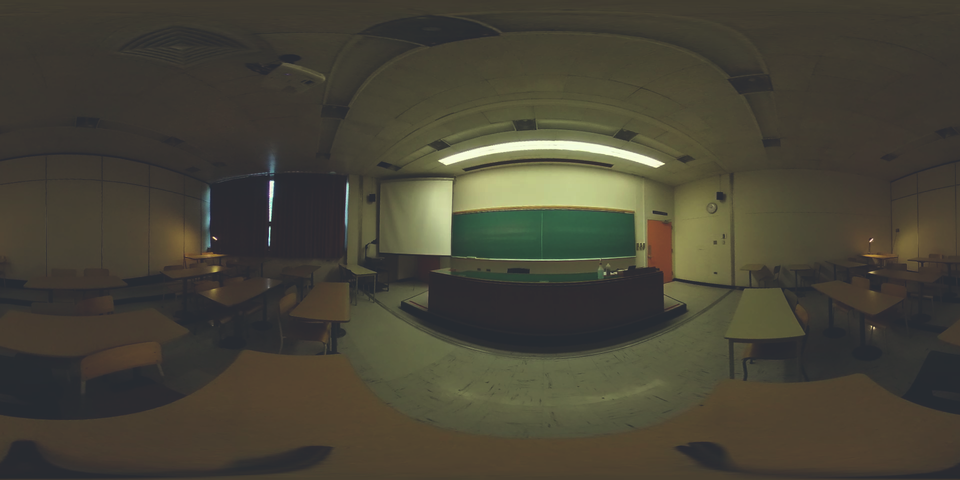} &
\includegraphics[width=0.248\linewidth]{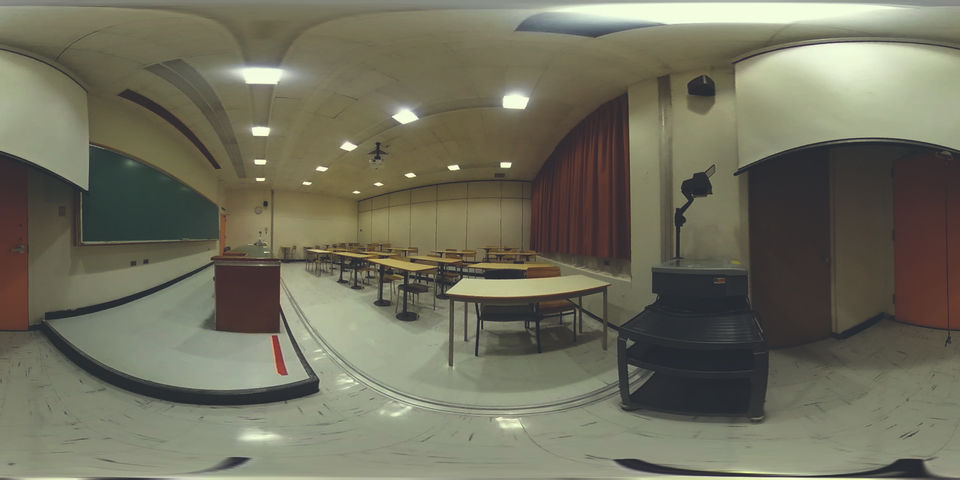}
\end{tabular}
\caption{Representative images from each test scene used in the experiments.}
\label{fig:dataset-mosaic}
\end{figure}



\noindent\textbf{Loss functions} \quad We train \methodname using supervision from LDR2HDR network. Traditional volume rendering methods are used to predict the radiance and densities of points sampled on the ray as in \cite{nerf++,mipnerf}. We define NeRF loss $\ell_\mathrm{nerf}$ between predicted radiance  $\hat{e}$ and ground truth HDR $e$ as 
\begin{equation}
    \label{eq:hdrnerf-loss}
    \ell_\mathrm{nerf} =  \sum_{r \in\mathcal{R}(\textbf{P})} \|\hat{E}(\textbf{r}) - E(\textbf{r})\|^2,
\end{equation}
where $E = log(1+e)$ and $\mathcal{R}(\textbf{P})$ is the set of camera rays at pose $\textbf{P}$.
The photographer, who inevitably is in the images, is segmented using an off-the-shelf segmentation algorithm~\cite{he2017mask}, and the corresponding pixels are ignored in the loss.

\noindent\textbf{Datasets} \quad To obtain training data for a given indoor scene, we casually capture a set of LDR panoramas using a commercial $360^\circ$ camera (Ricoh Theta Z1). We attach the camera to a portable tripod and capture a $360^\circ$ video while waving the camera around the scene to cover the entire volume as much as possible for a few minutes (typically 3--5 minutes for the scenes used in the experiments). Approximately 200 frames are then extracted from the video at even intervals. The camera parameters of the input LDR panoramas are recovered using OpenSFM\cite{opensfm} and given as input to \methodname. 

\begin{table}
\setlength{\tabcolsep}{2pt}
\fontsize{7}{9}\selectfont 
\centering

  \begin{tabular}{l|ccc|ccc|ccc}
    \toprule
      \multirow{3}{*}{Dataset} &
      \multicolumn{3}{c|}{Input LDR} &
      \multicolumn{3}{c|}{LDR2HDR pre-trained} &
      \multicolumn{3}{c}{LDR2HDR finetuned} \\
      & {PU-PSNR$\uparrow$} & {RMSE$\downarrow$} & {HDR-VDP3$\uparrow$} 
      & {PU-PSNR$\uparrow$} & {RMSE$\downarrow$} & {HDR-VDP3$\uparrow$} 
      & {PU-PSNR$\uparrow$} & {RMSE$\downarrow$} &  {HDR-VDP3$\uparrow$} \\
    \midrule
      Chess room & 31.659 & 0.051 & 8.067 & 33.994 & 0.048 & 8.234  & 36.995 & 0.005 & 8.492 \\
      Stairway & 31.964 & 0.224 & 7.881 & 33.297 & 0.213 & 8.016 & 33.685 & 0.019 & 8.489 \\
      Cafeteria & 25.299 & 5.378 & 6.098 & 26.664 & 5.268 & 6.418 & 28.499 & 4.061 & 7.164 \\
      Spotlights & 23.939 & 3.489 & 6.001 & 25.118 & 3.438 & 6.097 & 28.966 & 0.877 & 7.667 \\
      Dark class & 30.657 & 0.364 & 7.429 & 32.125 & 0.340 & 7.592 & 32.594 & 0.262 & 8.135 \\
      Small class & 32.353 & 2.162 & 8.018 & 34.095 & 1.889 & 8.267 & 35.465 & 0.221 & 8.678 \\
    \midrule
      Overall & 28.399 & 1.703 & 7.243 & 30.913 & 1.634 & 7.406 & \textbf{33.651} & \textbf{0.696} & \textbf{8.209}\\
    \bottomrule
   
  \end{tabular}
   \caption{Quantitative comparison of different strategies for recovering radiance across captured scenes. ``Input LDR'' are on the images captured by the camera, ``LDR2HDR pre-trained'' is on our network pre-trained on the Laval Indoor Dataset, and ``LDR2HDR finetuned'' is after the network finetuned to test camera. As expected, finetuning helps bridge the domain gap and significantly improves the results. Results are averaged over different viewpoints.}
   \label{tab:ldr2hdr-quant}
   \vspace*{-4mm}
\end{table}

\begin{figure*}[t]
    \centering
    \setlength{\tabcolsep}{1pt}
    \begin{tabular}{ccccc} 
        & \footnotesize Input LDR & \footnotesize LDR2HDR pre-trained & \footnotesize LDR2HDR finetuned & \footnotesize HDR ground truth \\ 
        \rotatebox[origin=c]{90}{\footnotesize Spotlights} &
        \includegraphics[width=0.23\linewidth,valign=m]{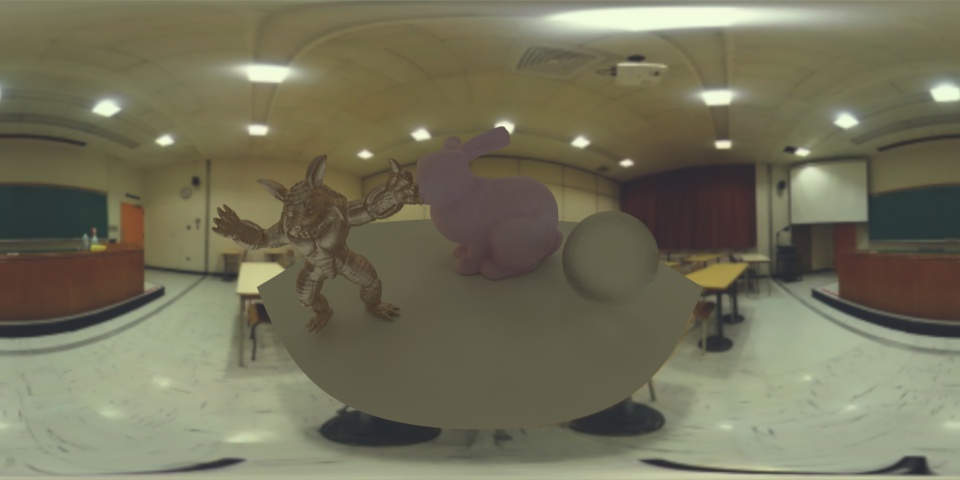} &
        \includegraphics[width=0.23\linewidth,valign=m]{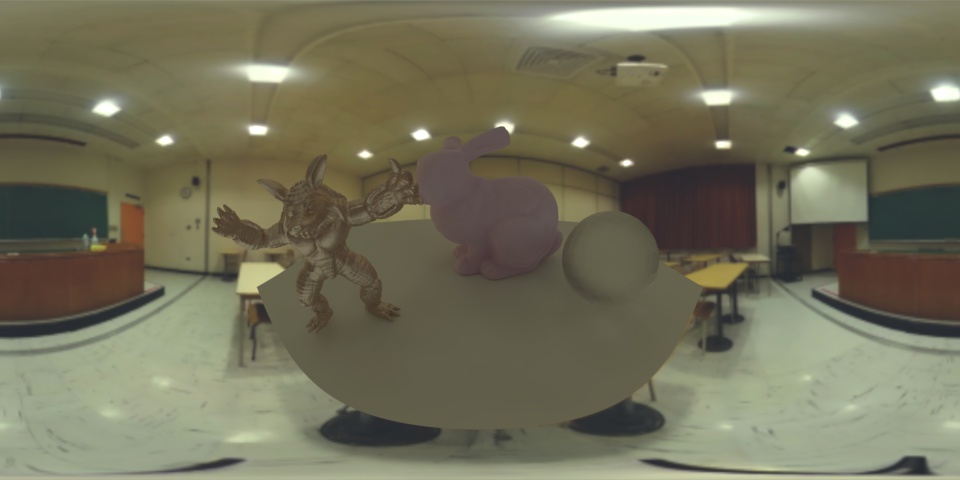} &
        \includegraphics[width=0.23\linewidth,valign=m]{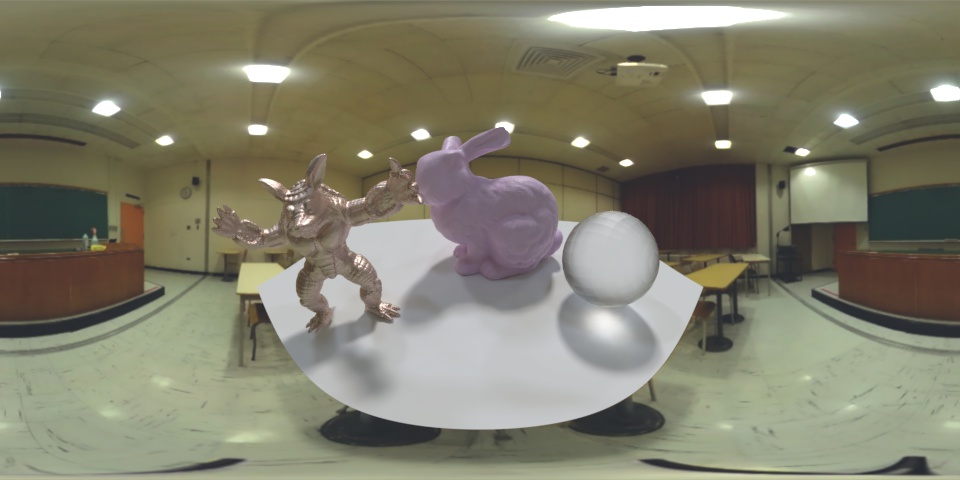} &
        \includegraphics[width=0.23\linewidth,valign=m]{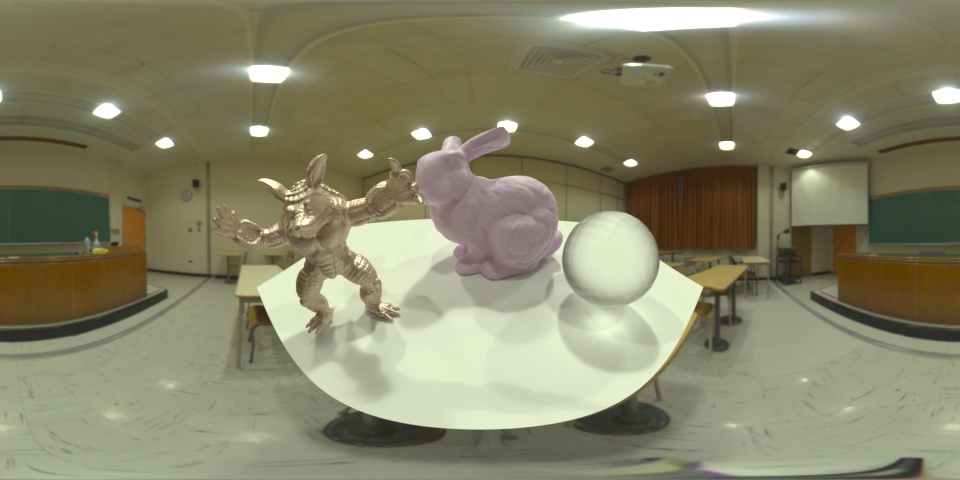} \\
        
        \rotatebox[origin=c]{90}{\footnotesize Stairway} &
        \includegraphics[width=0.23\linewidth,valign=m]{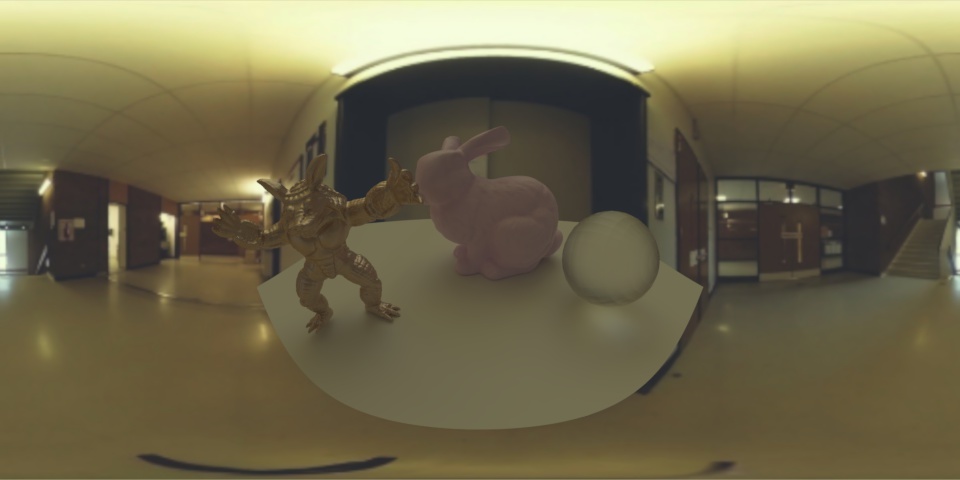} &
        \includegraphics[width=0.23\linewidth,valign=m]{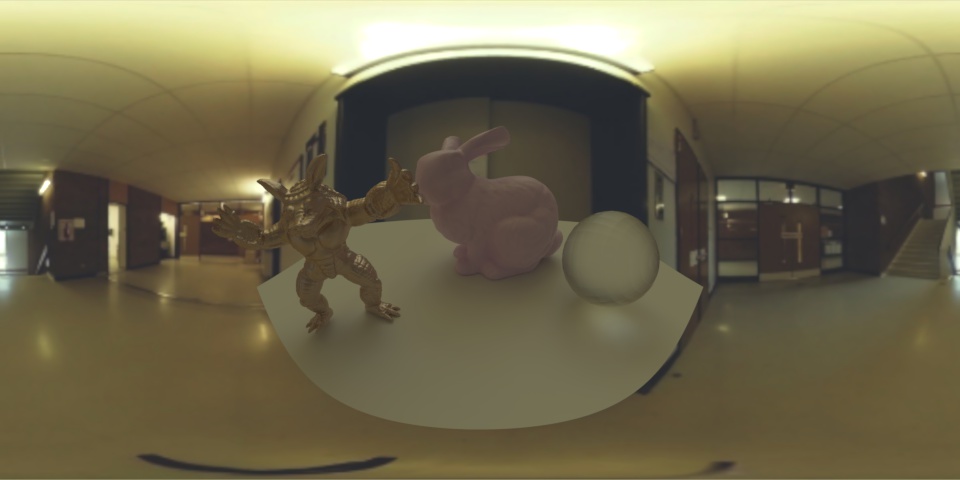} &
        \includegraphics[width=0.23\linewidth,valign=m]{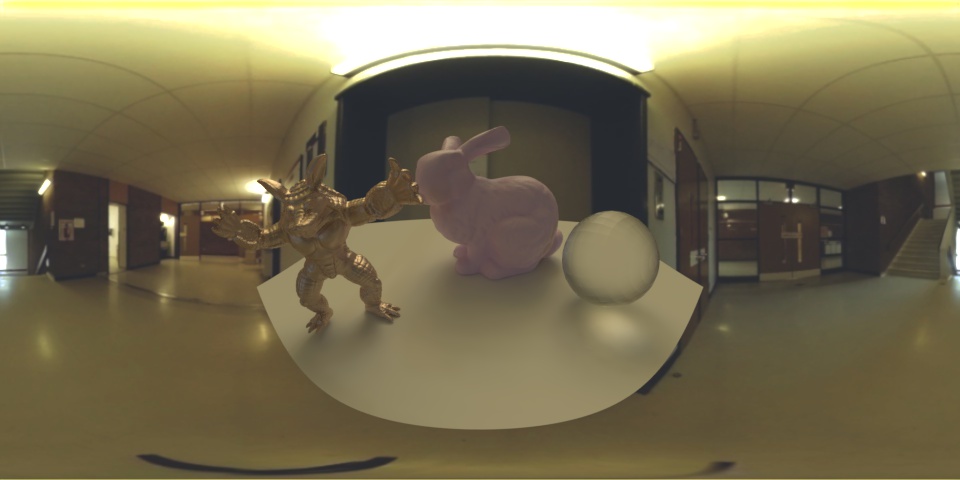} &
        \includegraphics[width=0.23\linewidth,valign=m]{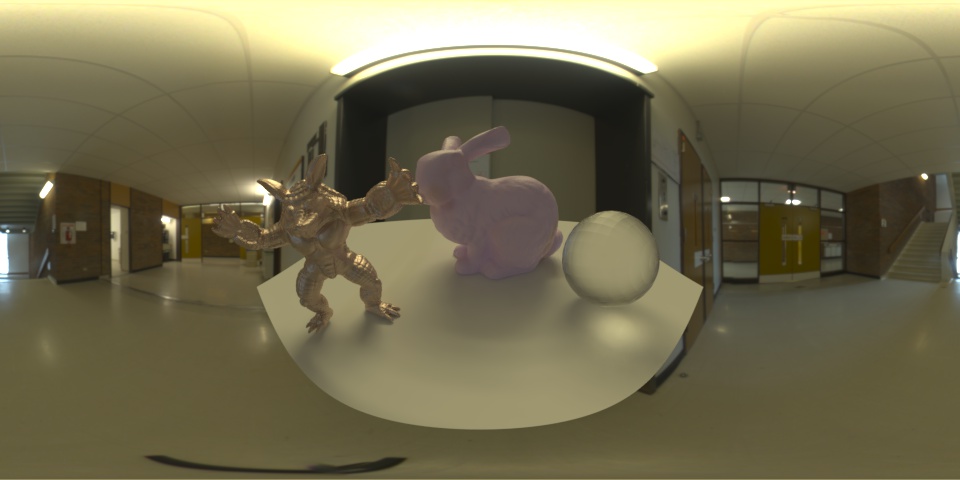} \\
        \rotatebox[origin=c]{90}{\footnotesize Small class} &
        \includegraphics[width=0.23\linewidth,valign=m]{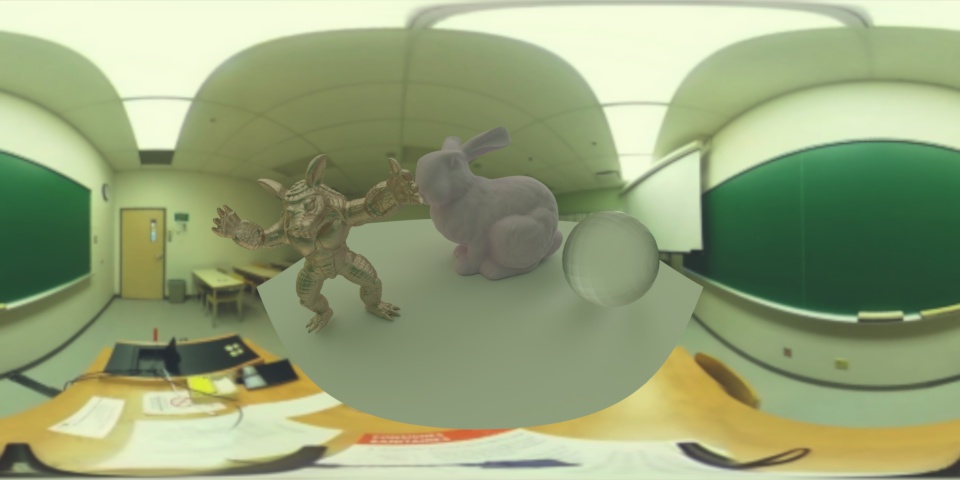} &
        \includegraphics[width=0.23\linewidth,valign=m]{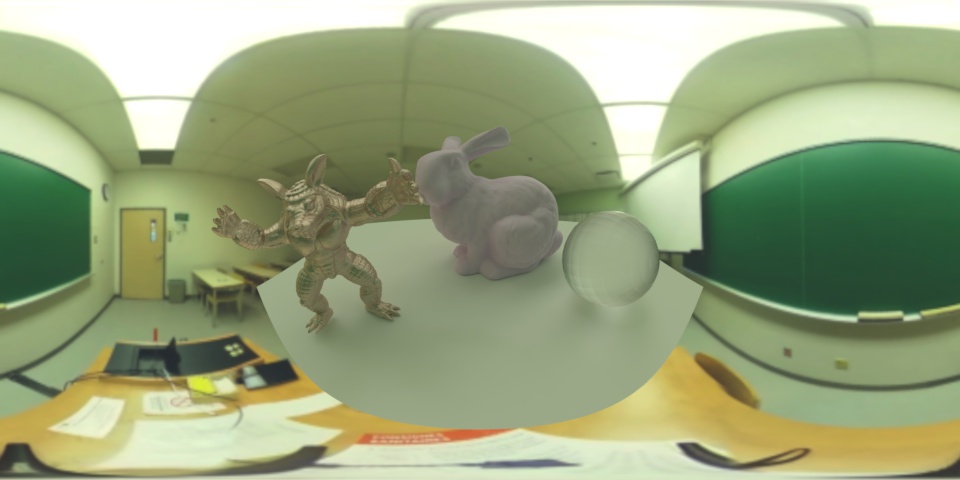} &
        \includegraphics[width=0.23\linewidth,valign=m]{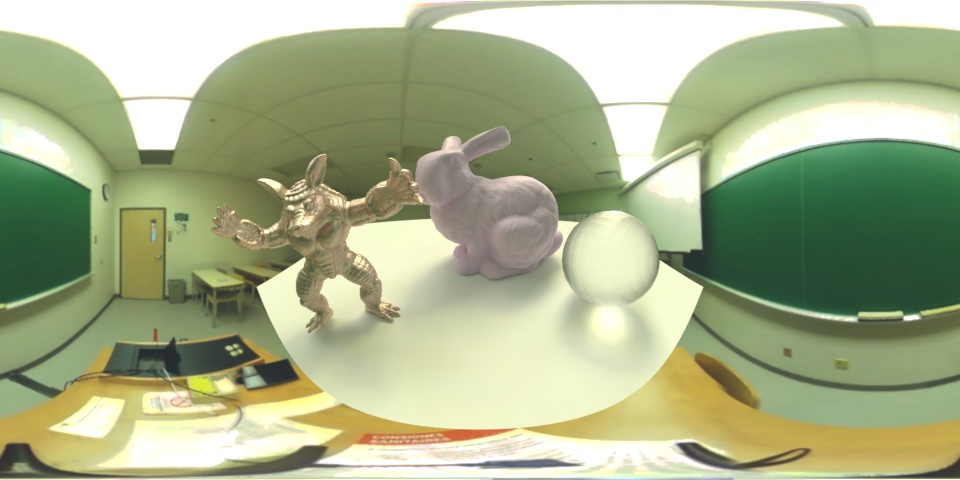} &
        \includegraphics[width=0.23\linewidth,valign=m]{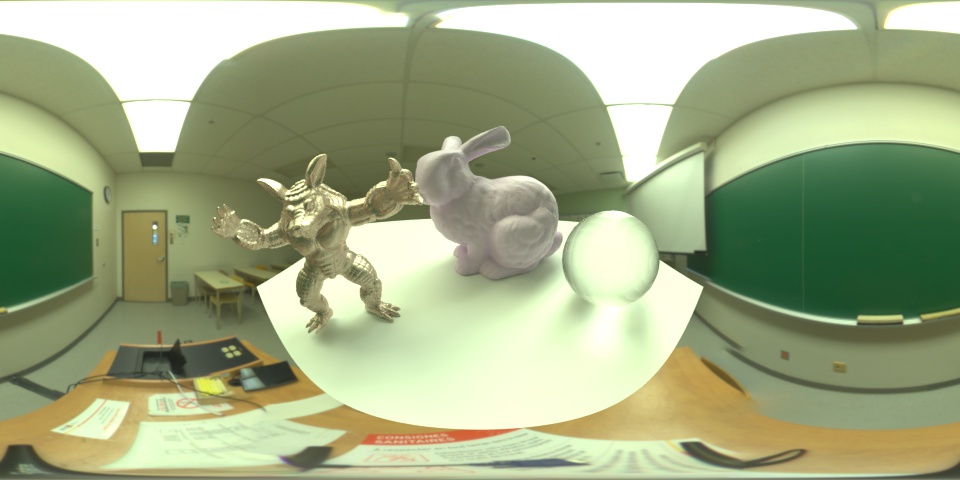} \\
    \end{tabular}
    \caption{Qualitative comparison of different strategies for recovering radiance across captured scenes. For each example, the figure shows virtual test objects relit to demonstrate the dynamic range. Note that despite some color imbalance (e.g. ``Spotlights''), fine-tuning helps bridge the domain gap between the training data and the captured images. Images tonemapped for display with $\gamma=2.2$.}
    \label{fig:ldr2hdr-qual}
\end{figure*}



\section{Evaluation}

In this section, we evaluate our approach against a set of challenging real scenes, where the ground truth HDR radiance is also captured at several locations. We further establish the sensitivity of design choices and compare them to closely related techniques. 

\subsection{Radiance capture evaluation dataset}
\label{sec:dataset}

For evaluation, we capture different real indoor scenes from a variety of different environments (\cref{fig:dataset-mosaic}). For each scene, we first capture a $360^\circ$ video using the Ricoh Theta Z1 panoramic camera as described in sec.~\ref{sec:panohdr-nerf}. We also capture a set of HDR panoramas for evaluation to use as ground truth. To obtain these, we set the camera on a tripod at certain locations in the scene (between 3--10 locations per scene) and program it using exposure bracketing to capture 11 exposures spanning over 22 f-stops that are subsequently merged to HDR using the PTGui Pro commercial software. We also capture a short video with the same camera parameters as the captured video at the same location as the HDR. We then extract a single frame from that video, allowing us to have an LDR image at the exact same location as its HDR counterpart. The resulting LDR image is linearized using a pre-calibrated camera response function obtained with a Macbeth color checker. In total, we capture six different scenes containing a total number of 10 LDR/HDR ground truth panorama pairs.



\subsection{LDR2HDR evaluation}
\label{sec:ldr2hdr-eval}

The LDR2HDR network presented in sec.~\ref{sec:ldr2hdr} is evaluated on the test set described in sec.~\ref{sec:dataset}. Tab.~\ref{tab:ldr2hdr-quant} compares the performance obtained by: using the LDR images directly, the LDR2HDR network pre-trained on the Laval Indoor HDR Dataset~\cite{Gardner2017LearningTP}, and after fine-tuning on the 78 HDR dataset captured with the same test camera. For evaluation, we use the ``PU-PSNR''~\cite{pu21}, which is a perceptually-uniform PSNR adjusted for HDR images. In addition, the ``RMSE'' corresponds to the rendering loss in \cref{eqn:loss-render}. Finally, we also report the HDR-VDP3~\cite{hdr-vdp}, where a value of 10 indicates a perfect match. Here, color encoding is set as ``rgb-bt.709'' for HDR evaluation, assuming a 24-inch display, $1920\times1080$ resolution, and a viewing distance of 1 meter.

As shown in tab.~\ref{tab:ldr2hdr-quant} and illustrated qualitatively in \cref{fig:ldr2hdr-qual}, there exists a significant domain gap between the training dataset and the test camera: simply pre-training on \cite{Gardner2017LearningTP} works marginally better than the input LDR image itself, but finetuning results in a significant performance gain on all metrics. Visually, finetuning produces renderings that look very similar to the ground truth (\cref{fig:ldr2hdr-qual}). 

\subsection{\methodname evaluation}
To evaluate how well \methodname works in terms of capturing the high dynamic range radiance field, we use the same set of ground truth HDR images as described in \cref{sec:dataset}. We infer environment maps at the locations of HDR panoramas and use them to render a synthetic scene. We modified NeRF++ to work with equirectangular image representation and trained directly on the LDR frames of the input video. The rendering results of NeRF++ appear dark compared to ground truth, and the lighting is not realistic (\cref{fig:nerf}). In addition, the generated shadows are soft and faded. In contrast, \methodname produces well-lit results, with sharp shadows that are similar to ground truth.
\subsection{Ablation study}

\begin{table}[t]
\setlength{\tabcolsep}{2pt}
\centering
\footnotesize

\label{tab:ldr2hdr-quant-laval}
\begin{tabular}{lcccc}
\toprule
& \multicolumn{2}{c}{w/o render loss} &  \multicolumn{2}{c}{render loss} \\
& PU-PSNR$\uparrow$ & HDR-VDP3$\uparrow$ & PU-PSNR$\uparrow$ & HDR-VDP3$\uparrow$ \\
\midrule
Hall. Net~\cite{hallu} & 30.06 & 7.54 & 31.51 & 8.12 \\
LANet~\cite{lanet} & 32.43 & 8.26 & \textbf{38.20} & \textbf{9.67} \\
\bottomrule
\end{tabular}
\caption{Quantitative comparison of two single image HDR estimation architectures on the Laval Indoor HDR test set, with and withough the rendering loss $\ell_\mathrm{render}$ while training the network. Render Loss with LANet significantly improves the results.}
\end{table}

\begin{table}[t]

\setlength{\tabcolsep}{2pt}
\fontsize{8}{9}\selectfont
\centering
  \label{tab:log-order}
  \begin{tabular}{lcccccc}
    \toprule
    \multirow{3}{*}{Dataset} &
    \multicolumn{2}{c}{Linear loss} &
    \multicolumn{2}{c}{\methodname} &
    \multicolumn{2}{c}{NeRF-LDR2HDR} \\
    & \footnotesize PU-PSNR$\uparrow$ & \footnotesize RMSE$\downarrow$ & \footnotesize PU-PSNR$\uparrow$ & \footnotesize RMSE$\downarrow$ & \footnotesize PU-PSNR$\uparrow$ & \footnotesize RMSE$\downarrow$ \\
    \midrule
    Chess room    & 35.152 & 0.011 & 36.941 & 0.012 & 35.991 & 0.006 \\
    Stairway      & 31.810 & 0.055 & 33.169 & 0.056 & 32.707 & 0.055 \\ 
    Cafeteria     & 24.139 & 4.376 & 28.029 & 4.179 & 26.537 & 5.298 \\ 
    Spotlights    & 26.719 & 0.909 & 28.657 & 0.431 & 27.324 & 1.619 \\
    Dark class    & 28.431 & 1.367 & 30.687 & 1.509 & 29.819 & 0.621 \\
    Small class   & 36.829 & 0.043 & 37.687 & 0.054 & 38.529 & 0.006 \\
    \midrule
    Overall & 29.725 & 1.071 & \textbf{32.528} & \textbf{1.038} & 31.650 & 1.301 \\
    \bottomrule
  \end{tabular}
  \caption{Quantitative comparison between linear (left) and log (middle) losses used for training. Comparison of \methodname with NeRF then LDR2HDR is at right }
\end{table}


\begin{figure*}
    \centering
    \footnotesize
    \newlength{\mywidth}
    \renewcommand{\mywidth}{0.19\linewidth}
    \setlength{\tabcolsep}{1pt}
    \begin{tabular}{ccccc} 
    Input LDR & NeRF++ & NeRF-LDR2HDR & \methodname (ours) & GT \\ 
      \includegraphics[width=\mywidth,valign=m]{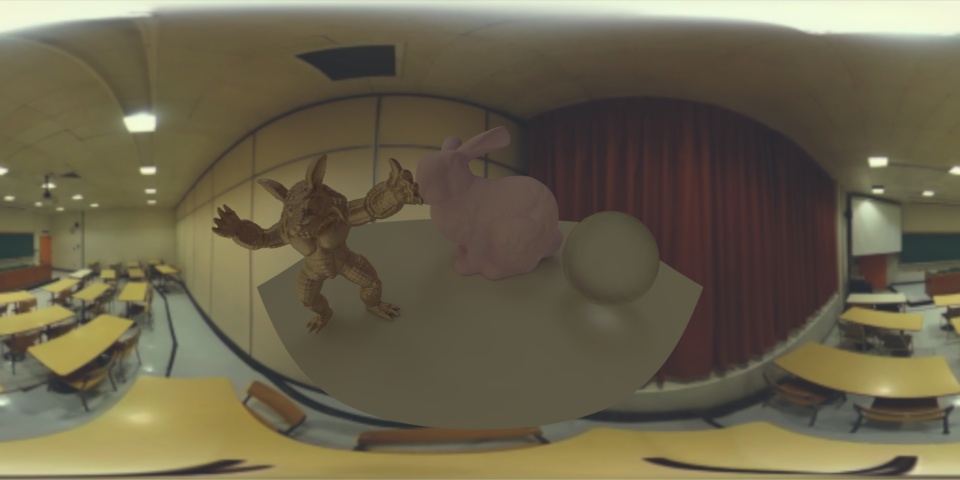} &
      \includegraphics[width=\mywidth,valign=m]{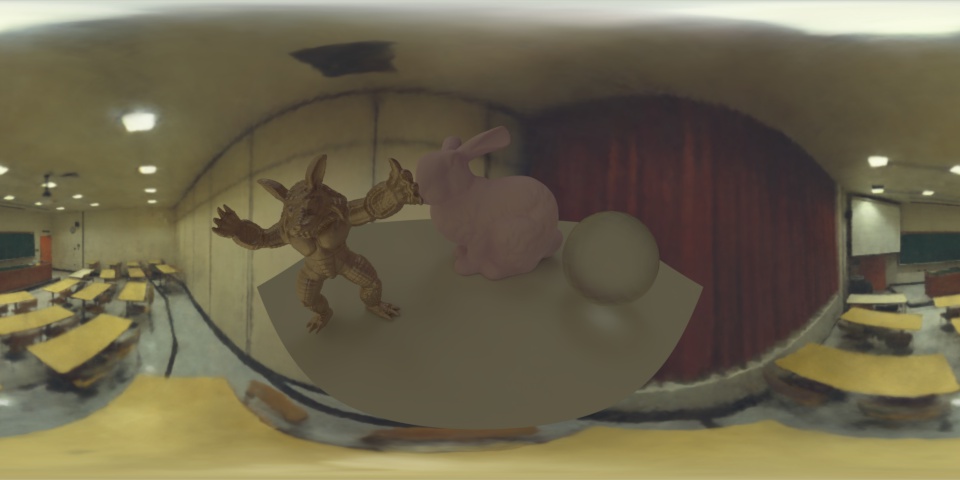} &
      \includegraphics[width=\mywidth,valign=m]{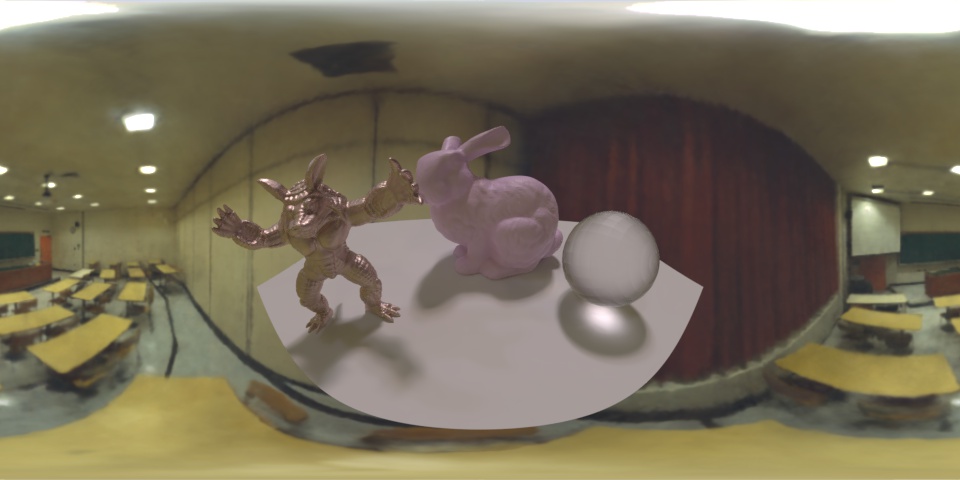} &
      \includegraphics[width=\mywidth,valign=m]{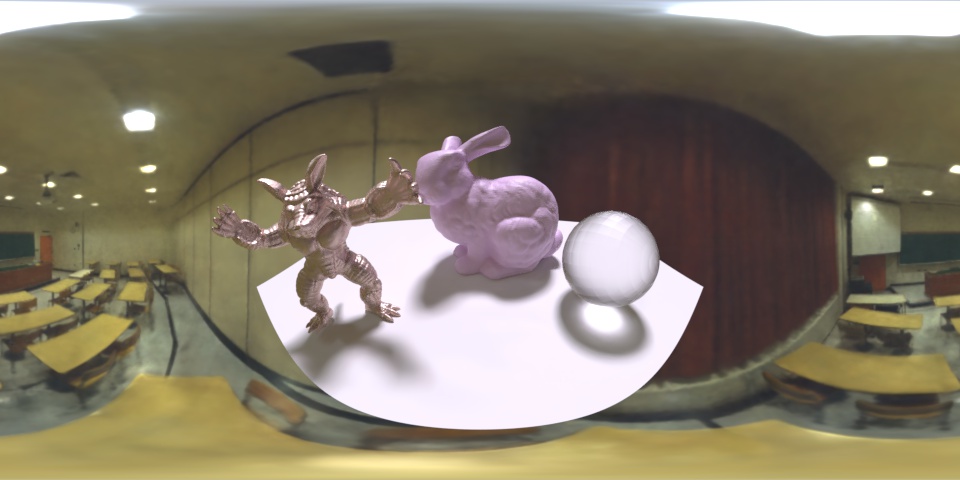} &
      \includegraphics[width=\mywidth,valign=m]{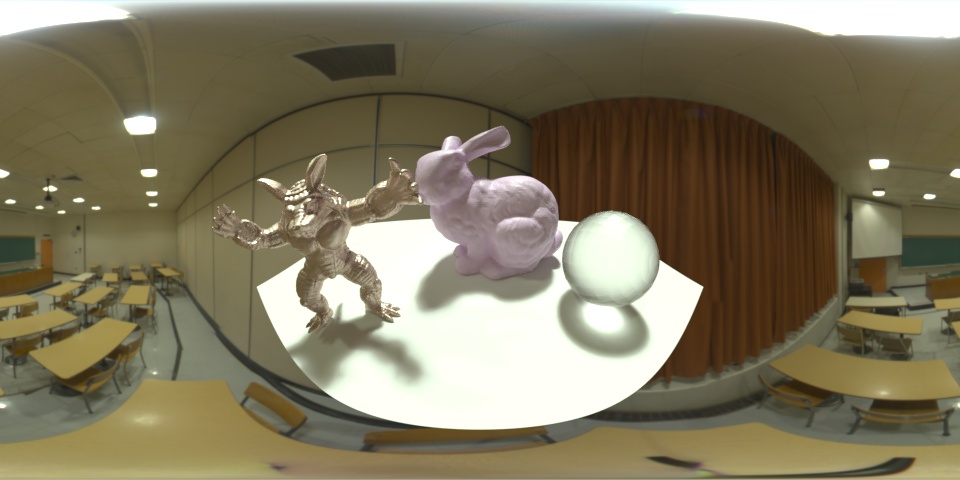} \\
       &&(a) Spotlights&&\\
      %
      \includegraphics[width=\mywidth,valign=m]{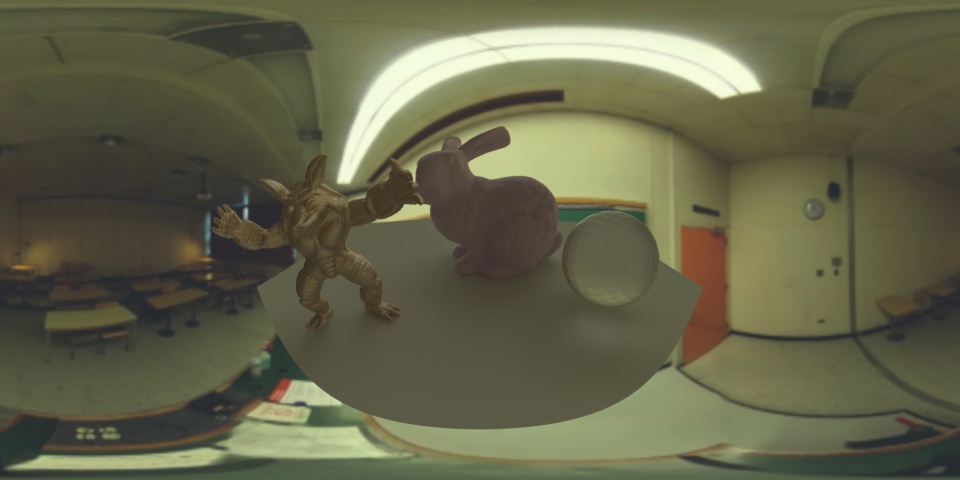} &
      \includegraphics[width=\mywidth,valign=m]{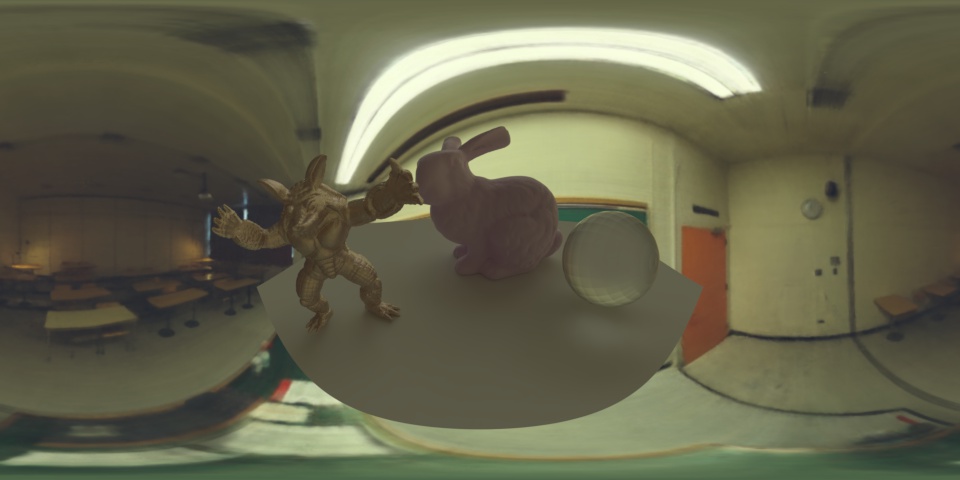} &
      \includegraphics[width=\mywidth,valign=m]{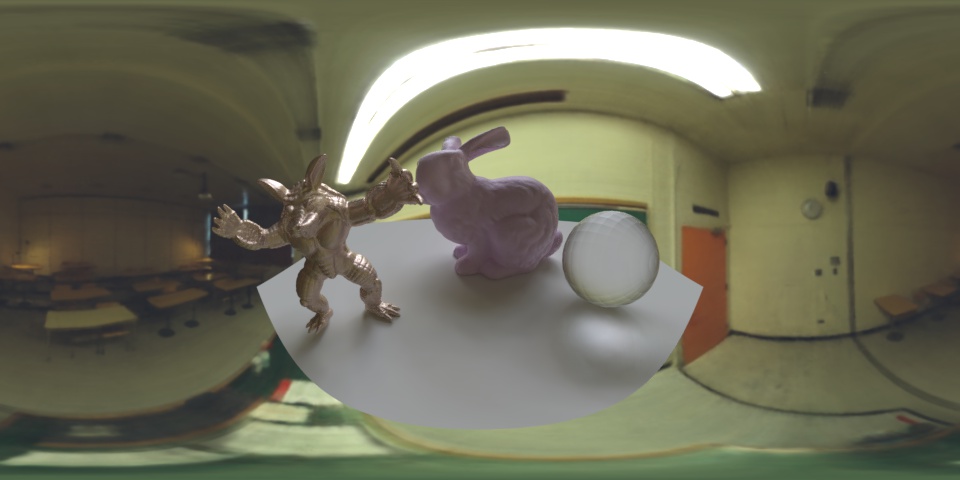} &
      \includegraphics[width=\mywidth,valign=m]{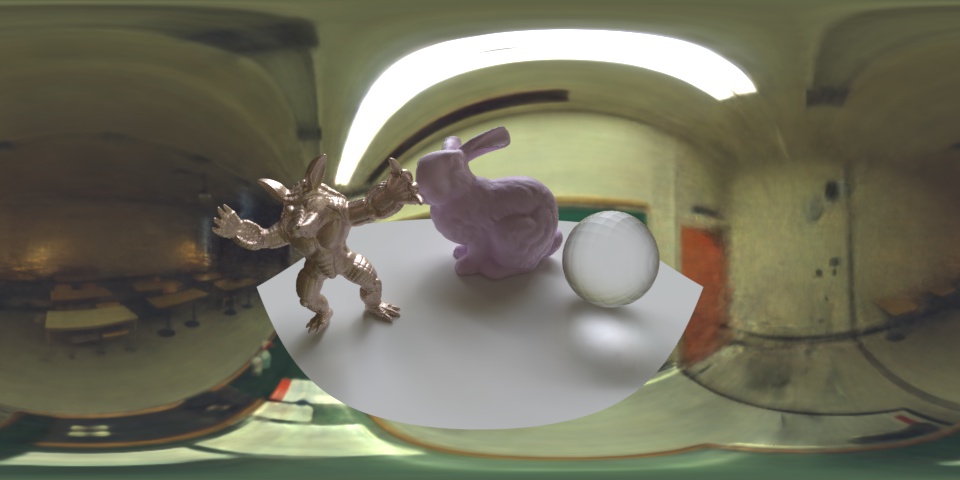} &
      \includegraphics[width=\mywidth,valign=m]{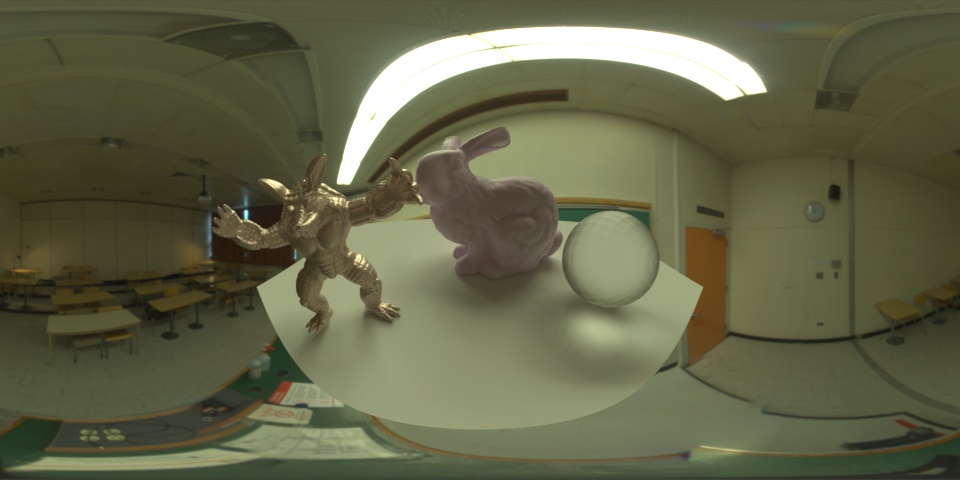} \\
      &&(b) Dark Class&&\\
      %
      \includegraphics[width=\mywidth,valign=m]{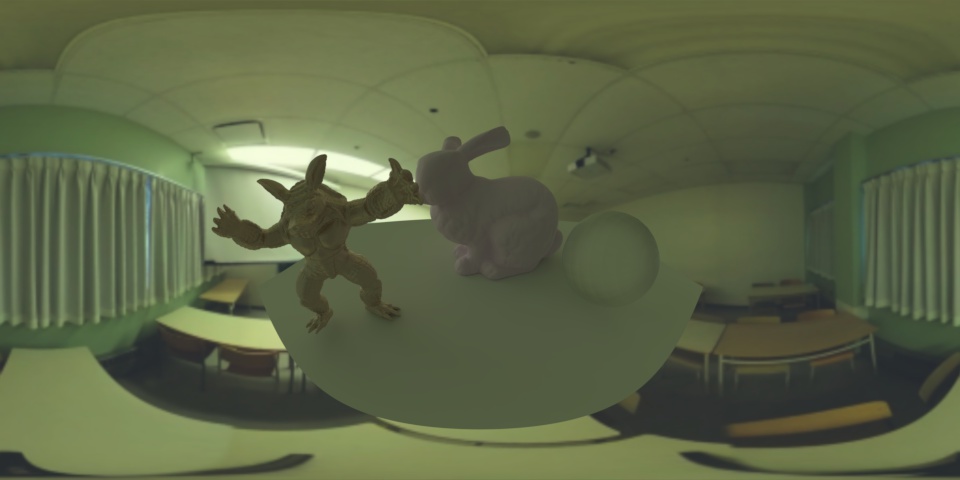} &
      \includegraphics[width=\mywidth,valign=m]{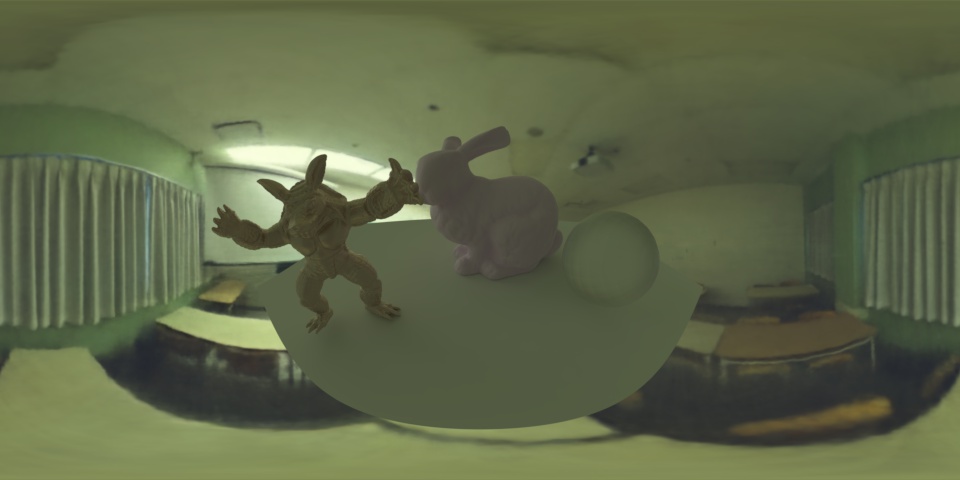} &
      \includegraphics[width=\mywidth,valign=m]{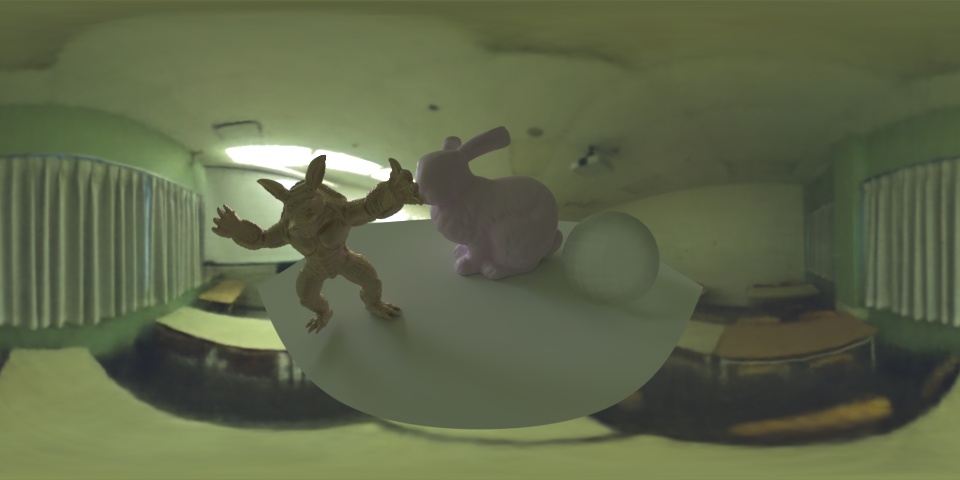} &
      \includegraphics[width=\mywidth,valign=m]{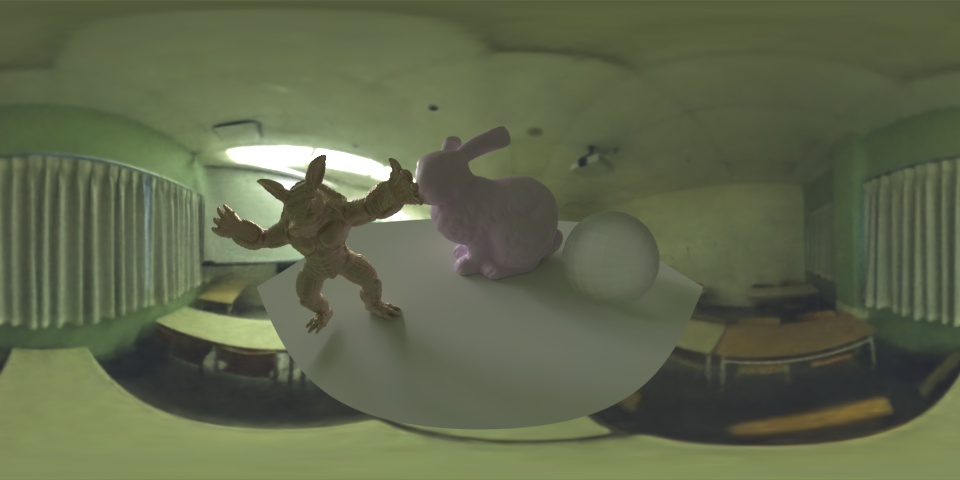} &
      \includegraphics[width=\mywidth,valign=m]{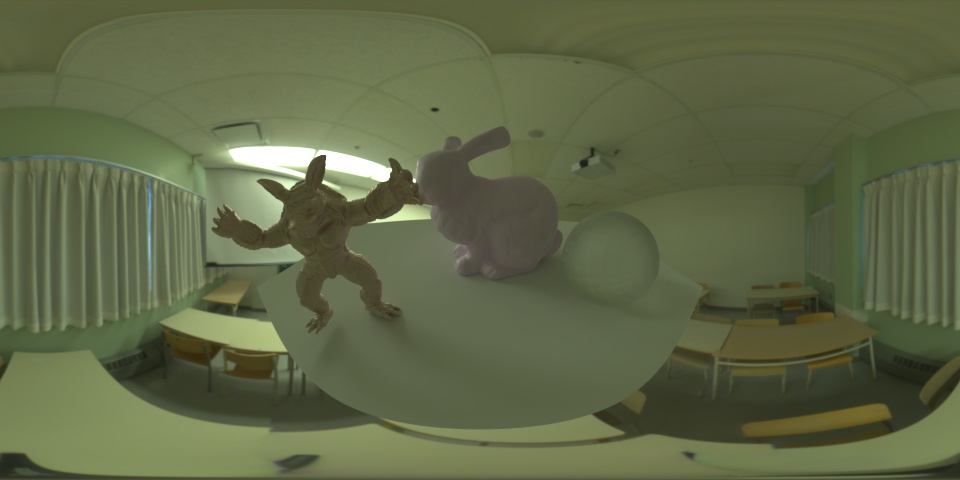} \\
      &&(c) Small Class&&\\
      %
      \includegraphics[width=\mywidth,valign=m]{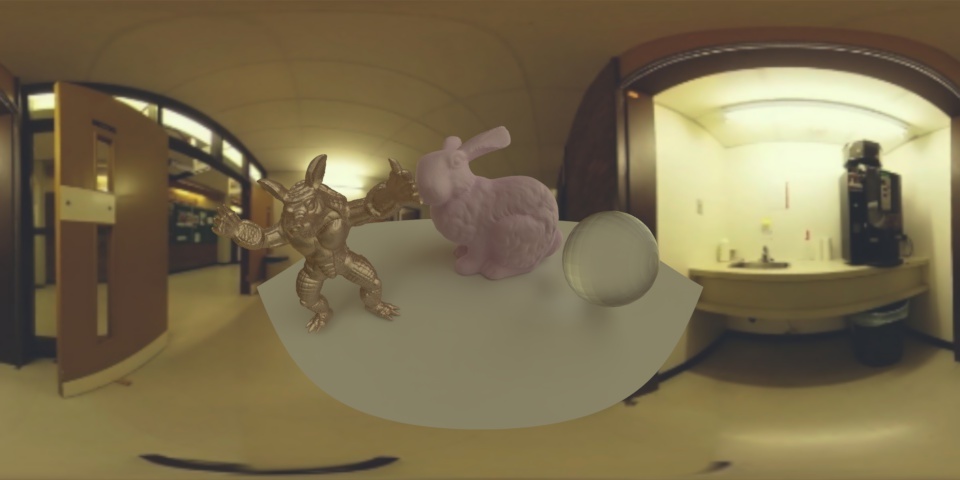} &
      \includegraphics[width=\mywidth,valign=m]{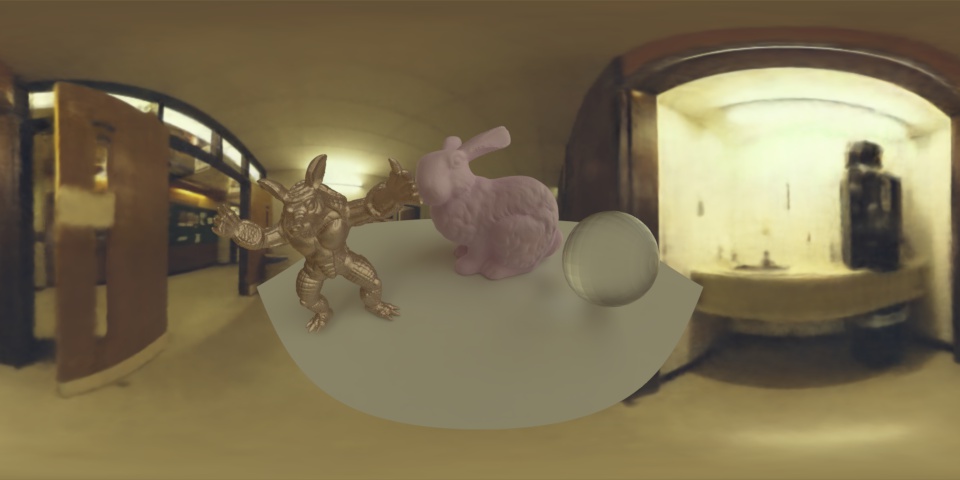} &
      \includegraphics[width=\mywidth,valign=m]{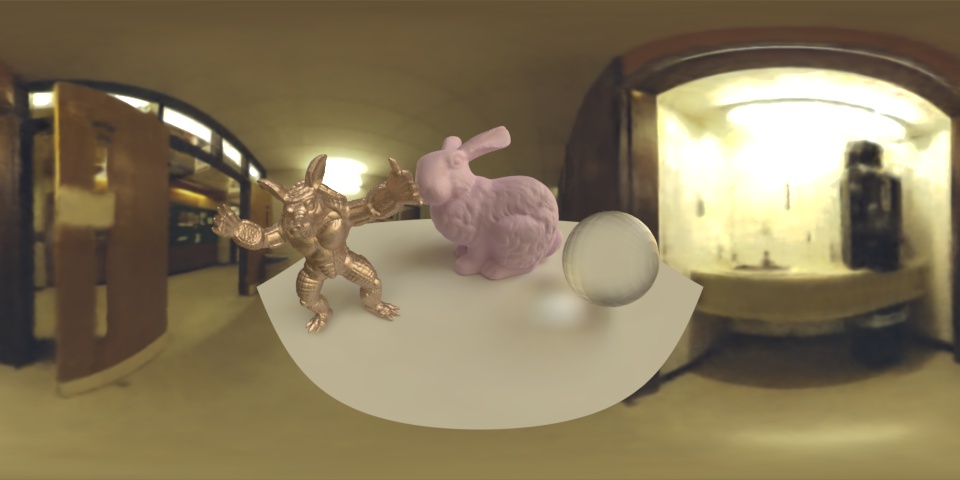} &
      \includegraphics[width=\mywidth,valign=m]{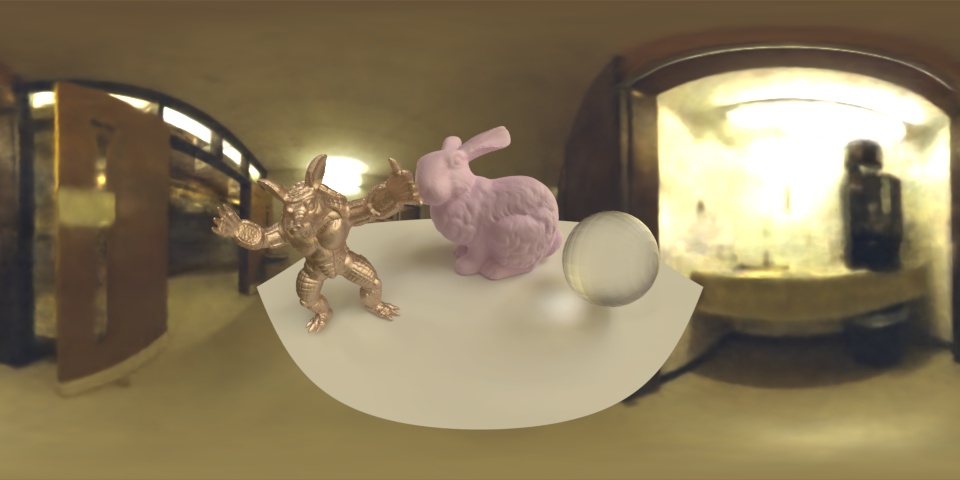} &
      \includegraphics[width=\mywidth,valign=m]{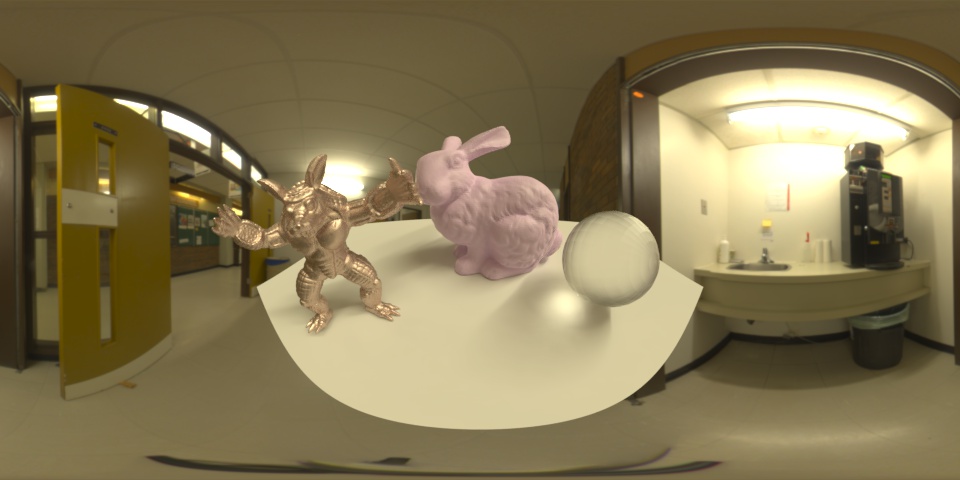} \\
      &&(d) Stairway&&\\
      %
      \includegraphics[width=\mywidth,valign=m]{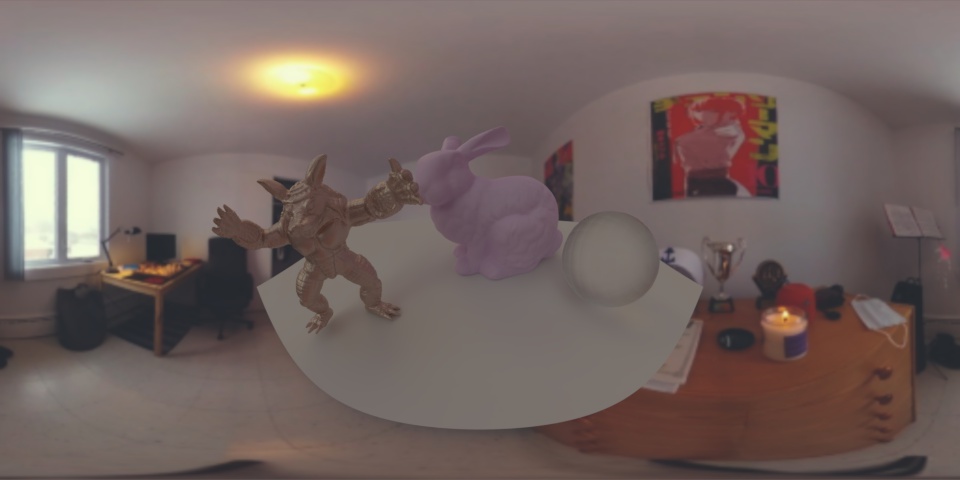} &
      \includegraphics[width=\mywidth,valign=m]{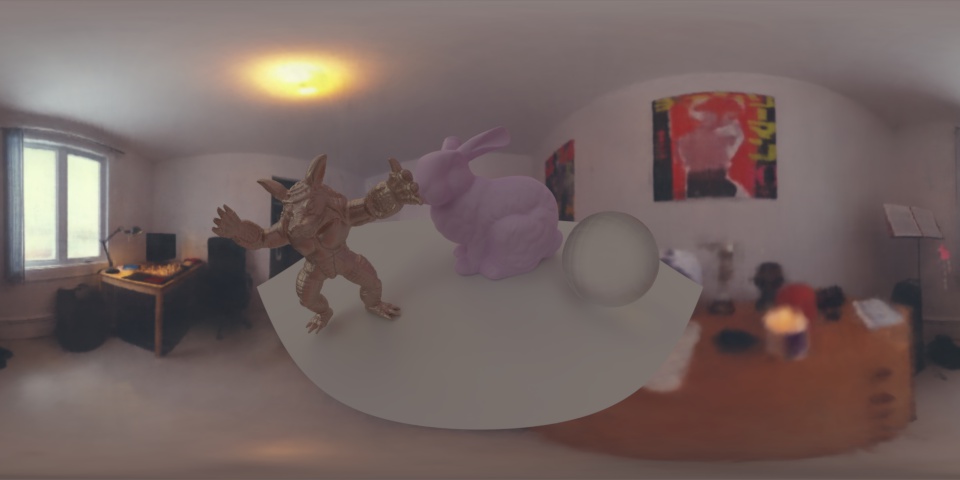} &
      \includegraphics[width=\mywidth,valign=m]{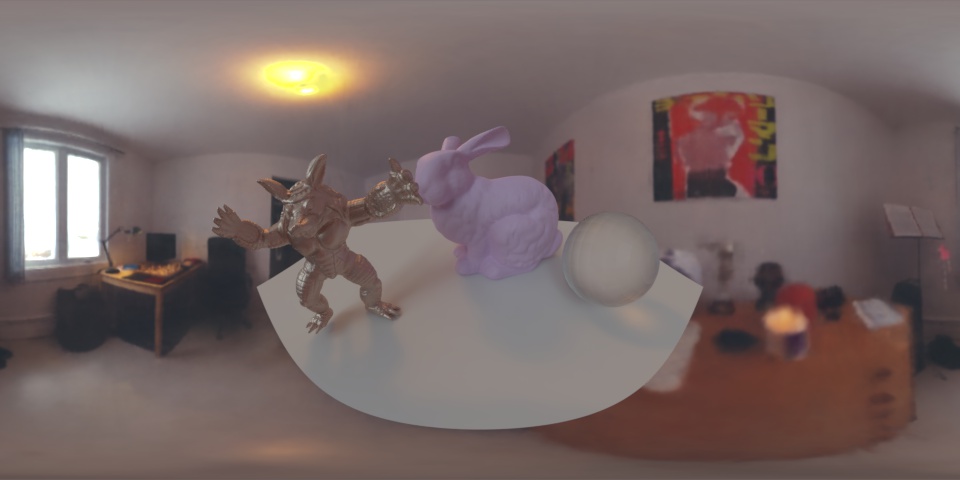} &
      \includegraphics[width=\mywidth,valign=m]{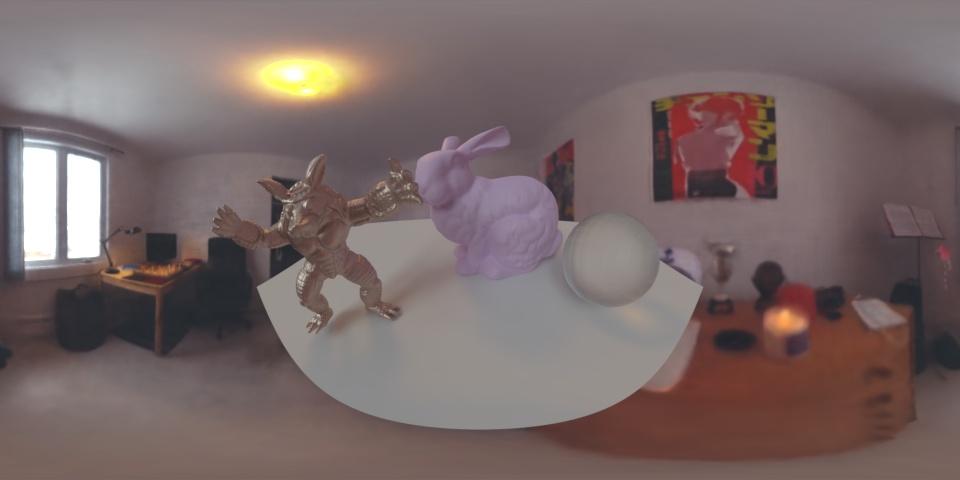} &
      \includegraphics[width=\mywidth,valign=m]{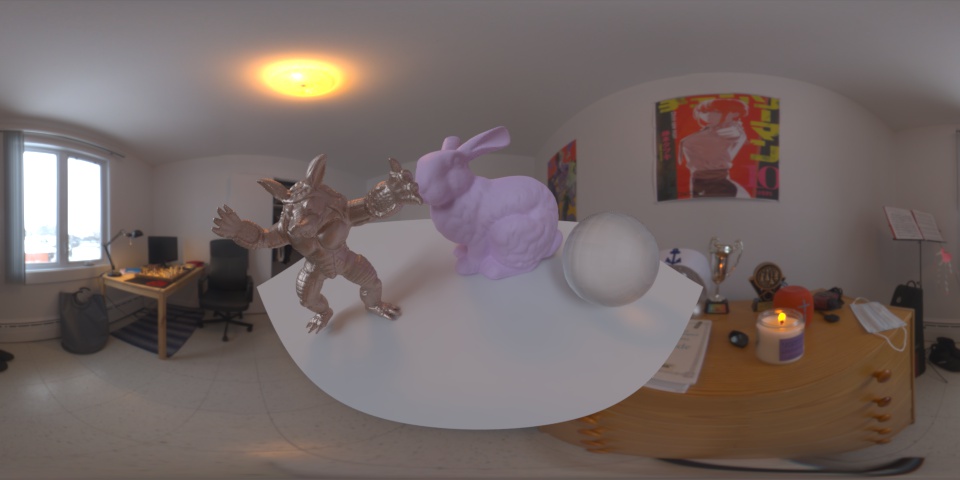} \\
      &&(d) Chess Room&&\\
      %
      \includegraphics[width=\mywidth,valign=m]{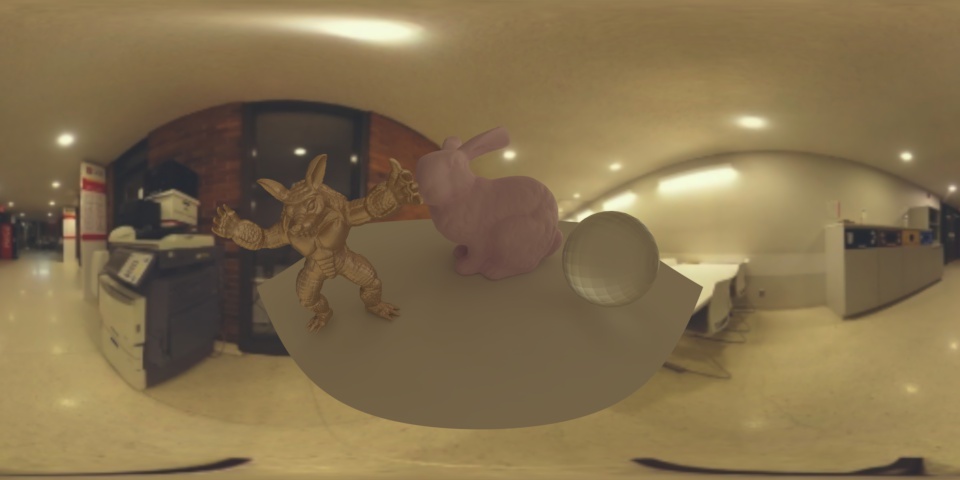} &
      \includegraphics[width=\mywidth,valign=m]{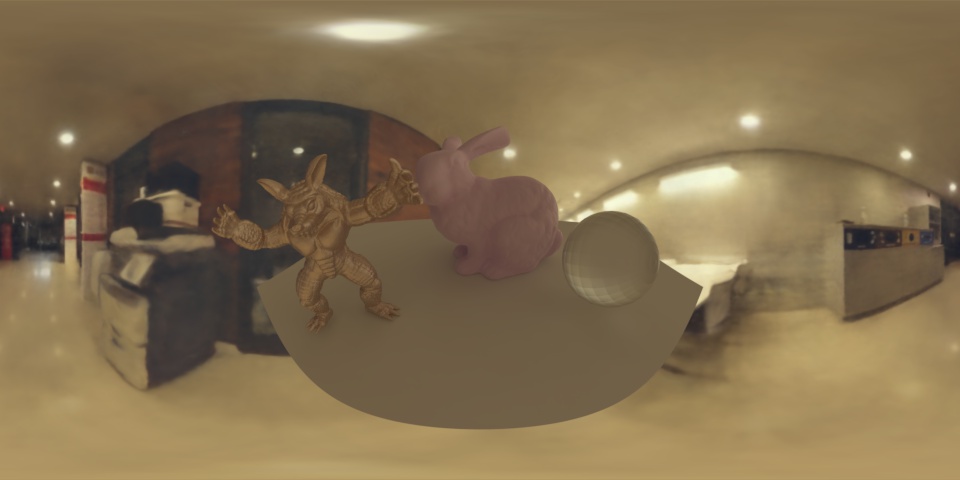} &
      \includegraphics[width=\mywidth,valign=m]{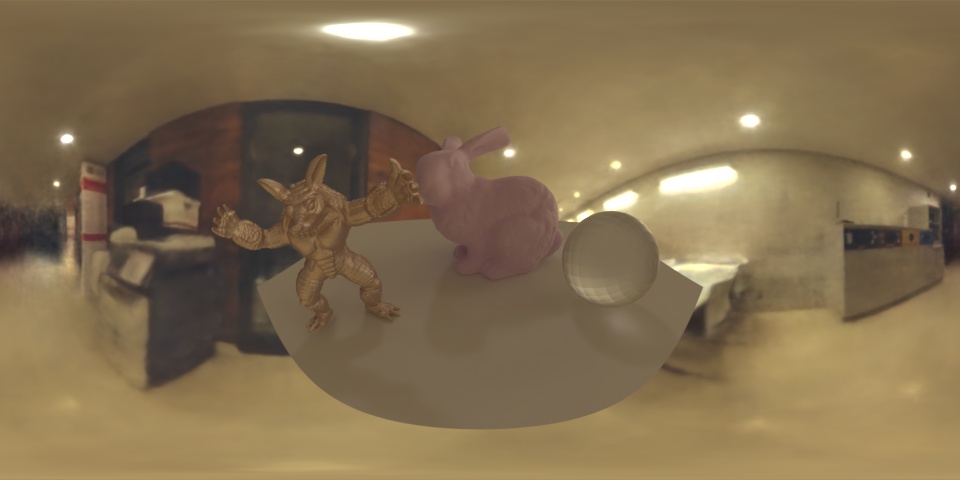} &
      \includegraphics[width=\mywidth,valign=m]{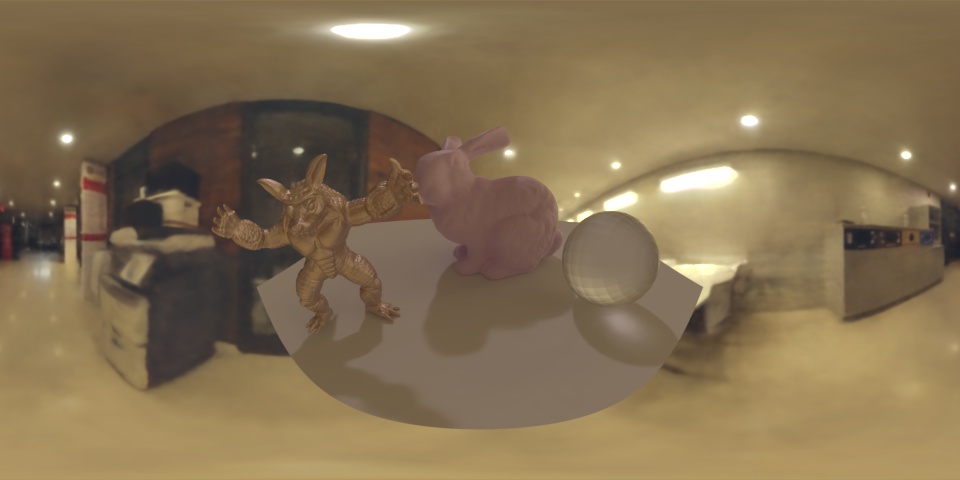} &
      \includegraphics[width=\mywidth,valign=m]{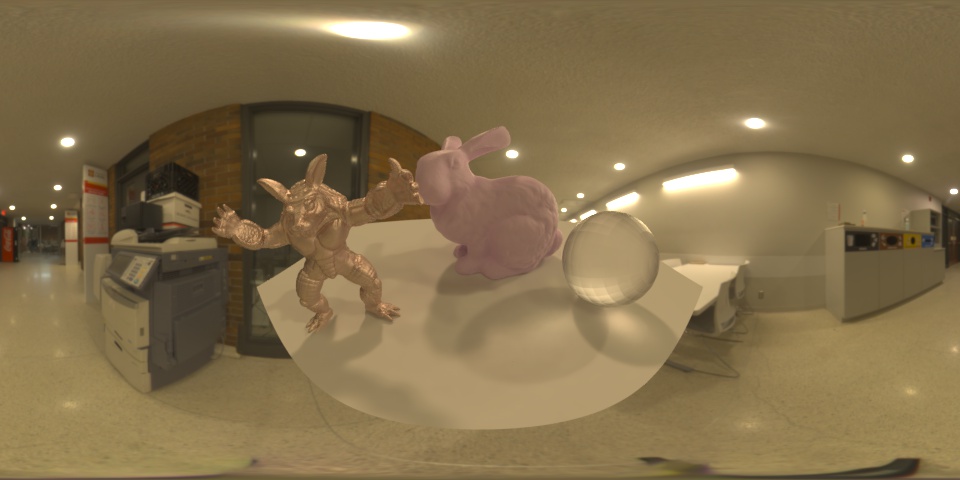} \\
      &&(e) Cafeteria&&
    \end{tabular}
    \caption{Comparing input LDR, NeRF++, NeRF-LDR2HDR, \methodname (ours), and GT panoramas. For each example, the figure shows a virtual test object relit to show the dynamic range. While NeRF++ is able to model the scene correctly, it is unable to capture the radiance of the scene. \methodname is able to faithfully capture the radiance of the scene. We compare it with NeRF-LDR2HDR which estimates HDR from NeRF++ outputs. Although it is able to closely estimate the radiance, it leads to flickering between consecutive frames. Images tonemapped for display with $\gamma=2.2$.}
    \label{fig:nerf}
    \vspace*{-5mm}
\end{figure*}
\begin{figure}
    \centering
    \footnotesize
    \setlength{\tabcolsep}{1pt}
    \begin{tabular}{cccc} 
        & Linear space & Log space & HDR ground truth \\ 
        \rotatebox[origin=c]{90}{Spotlights} &
        \includegraphics[width=0.2\linewidth,valign=m]{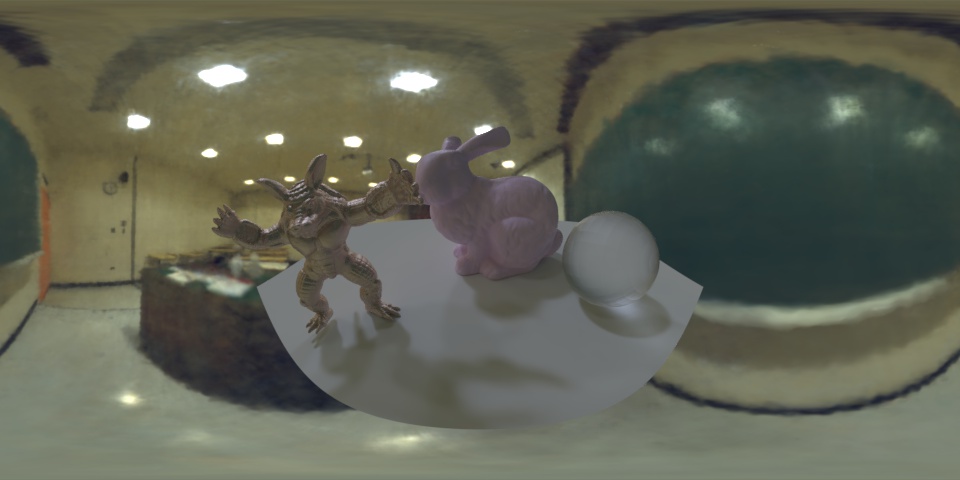} &
        \includegraphics[width=0.2\linewidth,valign=m]{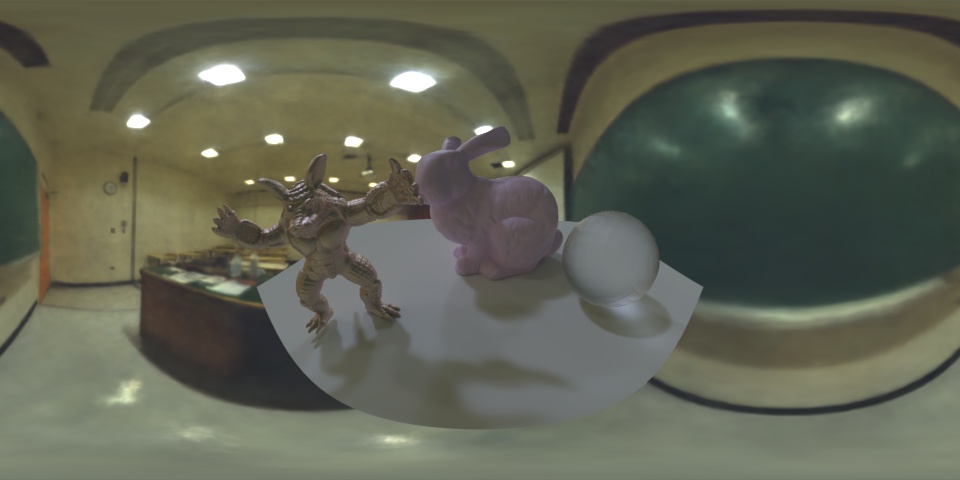} &
        \includegraphics[width=0.2\linewidth,valign=m]{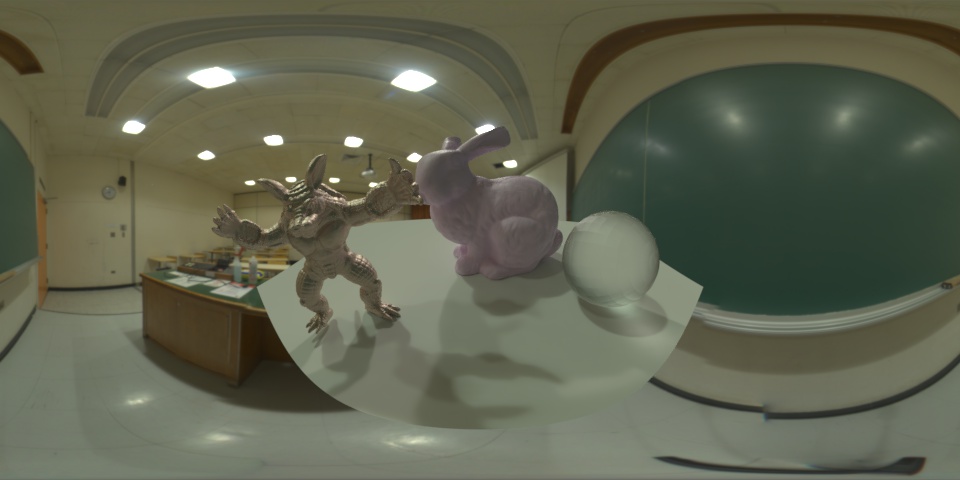} \\

        \rotatebox[origin=c]{90}{Stairway} &
        \includegraphics[width=0.2\linewidth,valign=m]{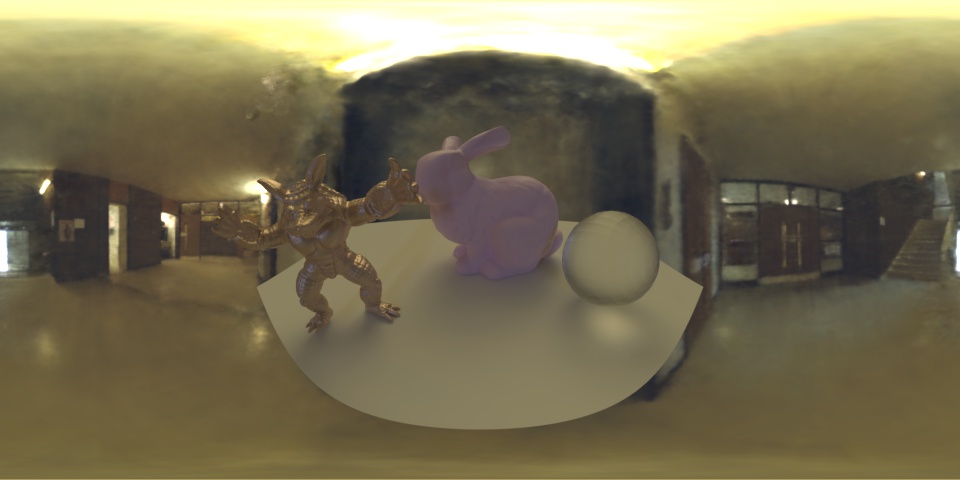} &
        \includegraphics[width=0.2\linewidth,valign=m]{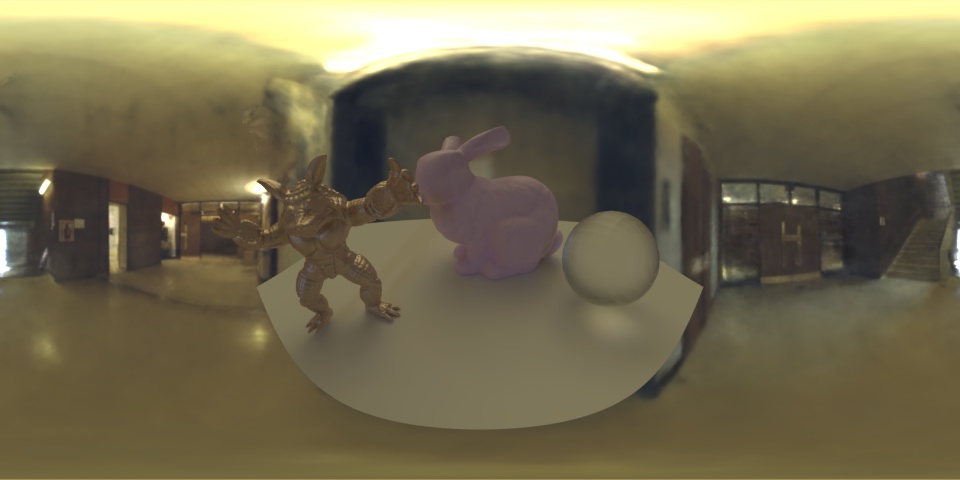} &
        \includegraphics[width=0.2\linewidth,valign=m]{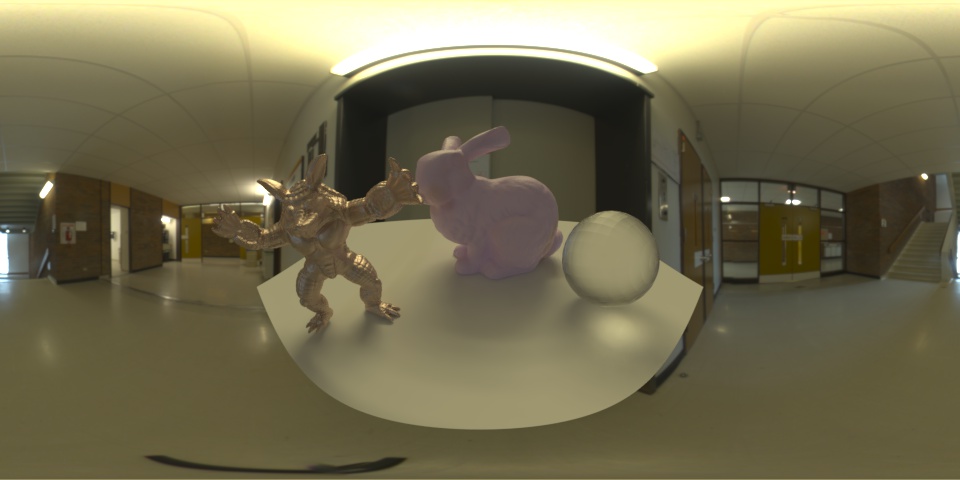} \\

        \rotatebox[origin=c]{90}{Chess room} &
        \includegraphics[width=0.2\linewidth,valign=m]{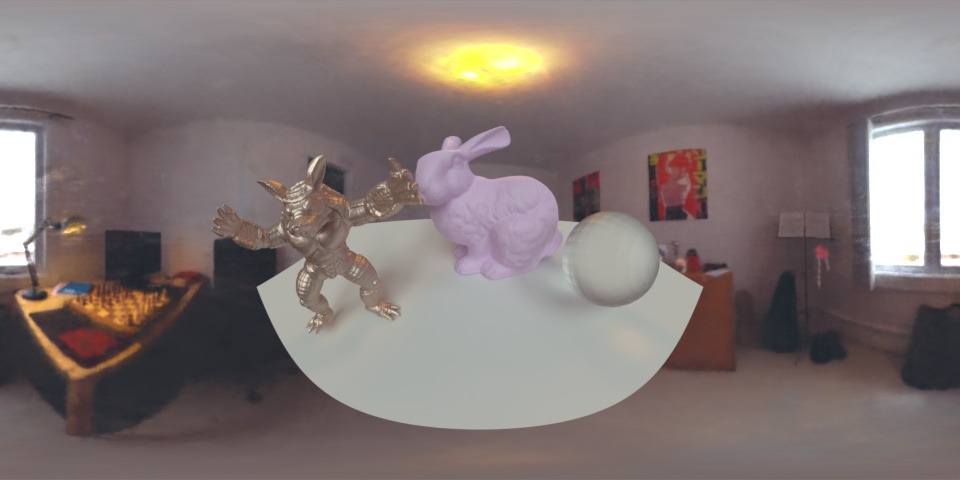} &
        \includegraphics[width=0.2\linewidth,valign=m]{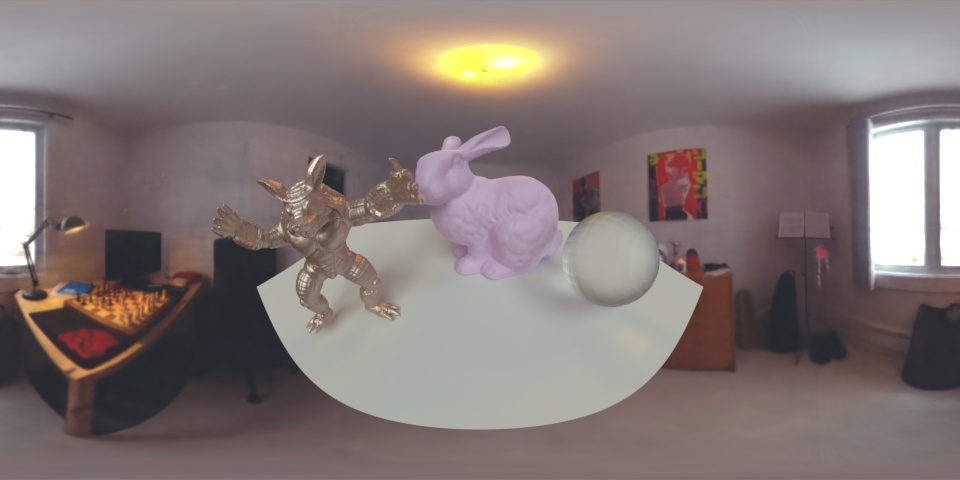} &
        \includegraphics[width=0.2\linewidth,valign=m]{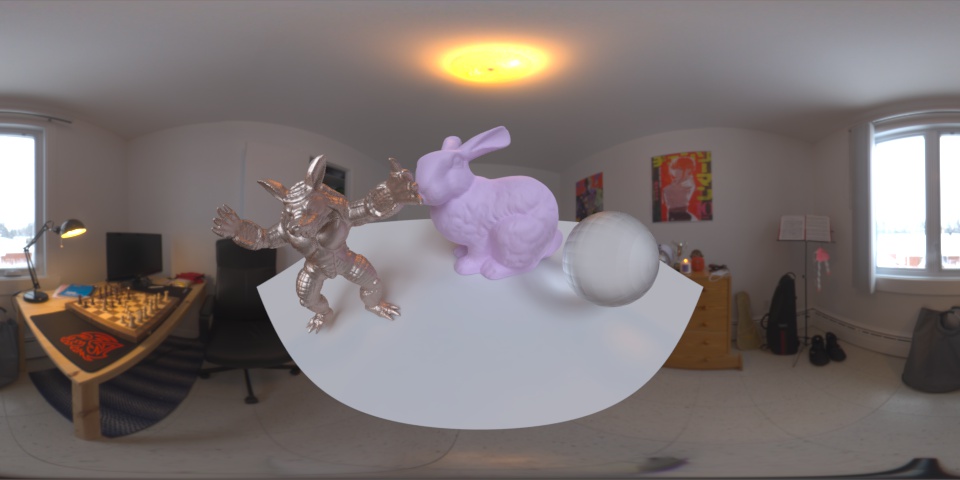} \\
    \end{tabular}
    \caption[]{Comparing panoramas generated by \methodname after loss in linear space and log space. For each example, the figure shows (top) the panorama with (bottom) virtual test objects relit to show the dynamic range. While the network learns the high dynamic range in both cases, we observe that taking loss in linear space leads to poor visual quality and floating artifacts in the output panoramas. PanoHDR-NeRF produces better results when trained in log space, consistent with \cite{lanet,deephdr,hdrnet}.}
    \label{fig:log_loss}
\end{figure}

\noindent\textbf{LDR2HDR network architecture} \quad  We compare between two recent SOTA single image HDR estimation architectures, namely HallucinationNet~\cite{hallu} and LANet~\cite{lanet}, with and without our rendering loss. For this, we evaluate on the Laval Indoor HDR~\cite{Gardner2017LearningTP} test set 
and report results in \cref{tab:ldr2hdr-quant-laval}. With its saturation-driven attention, the LANet architecture outperforms HallucinationNet. In addition, the use of the rendering loss $\ell_\mathrm{rend}$ (\cref{eqn:loss-render}) helps the network focus on the bright light sources, which is crucial for accurate radiance reproduction.

\noindent\textbf{Planar vs spherical sampling} \quad We compare the impact of planar vs spherical sampling (check supplementary) for training our network on LDR images. In both cases, we follow the hierarchical sampling strategy of NeRF~\cite{nerf} and train the network with 64 coarse samples and 128 fine samples. We observe that sphere sampling outperforms planar sampling for omnidirectional images as it does not oversample points at the poles but does so uniformly on the sphere. (more details in supplementary)

\noindent\textbf{Loss in log space} \quad We evaluate the importance of computing the loss in log space (c.f. \cref{sec:panohdr-nerf}) in \cref{tab:log-order}, which suggests that the network is able to estimate the high dynamic range equally well with or without the log-space loss. However, \cref{fig:log_loss} shows that training in linear space results in more floating artifacts and blurrier images than those obtained by the log loss. This is because HDR intensities vary significantly and log space helps us restrict the range which in turn makes it easier for NeRF-MLP to predict the radiance values.

\noindent\textbf{Order of operations} \quad \methodname uses a NeRF network trained on HDR images. It is also possible to reverse the order by training the NeRF on LDR images and pass its output through the LDR2HDR network (dubbed ``NeRF-LDR2HDR''). We compare these options in \cref{tab:log-order} and \cref{fig:nerf}. Though the metrics given in the table don't differ much, the supplementary video shows \methodname produces temporally more stable results. This could be due to the averaging that naturally happens within the NeRF network. 

\section{Discussion and conclusion}


The main contribution of this work is a novel pipeline to predict the full HDR radiance of an indoor scene without using special hardware, careful scanning of the scene, or intricately calibrated camera configurations. Our pipeline can work with a single, off-the-shelf $360^\circ$ camera that is moved around the scene. Although \cite{rawnerf,hdrnerf} have demonstrated recovery of HDR intensities of forward facing scenes, they require multiple sets of LDR images at various exposures. Recovering HDR intensities for large indoor scenes using these methods requires an elaborate setup which is tedious and cumbersome. \methodname can render novel $360^\circ$ views from any point within an unbounded indoor scene in high dynamic range from a casually captured video. We show their use for the realistic relighting of virtual objects in real scenes, hopefully getting one step closer to democratizing augmented reality.


\noindent\textbf{Limitations and future work} \quad
Blurriness of the NeRF results is a big limitation of this work, despite using cone-casting from Mip-NeRF. We believe further improvements such as \cite{barron2022mipnerf360} can help in reconstructing sharper estimates.
Another limitation is that the photographer capturing the scene ends up modifying the light field ever so slightly by casting shadows, creating reflections off of shiny surfaces, etc. Unfortunately, the intensity changes this creates are too soft for existing shadow detectors~\cite{shadow}. Methods modeling transient changes~\cite{nerfw} could potentially be of help. The finetuning for each camera introduces additional effort though only once per camera. Future research can improve generalization across cameras, perhaps using multiple cameras for training, or through other data augmentation techniques.
Our approach learns radiance and view synthesis in two independent steps by specialized networks. Exploring how both can be done simultaneously, potentially in conjunction with geometry and material estimation~\cite{nerd,nerfactor}, is an exciting direction for future work. \methodname assumes that the scene is static, so the lighting should remain unchanged for the inference time. In addition, \methodname is an offline method and therefore the novelty of the proposed method is limited to only capturing process and not rendering. Recent efforts have demonstrated how to significantly shorten training~\cite{instantngp,yu2021plenoctrees} and inference~\cite{kilonerf,fastnerf} times of NeRF-based approaches, these could straightforwardly be incorporated into our framework. By reducing the time between capture and visualization, \methodname can be used for AR/VR applications such as virtual tours and VFX generation.
 



\section{Acknowledgments}
This research was supported by NSERC grant RGPIN-2020-04799, Compute Canada, and a MITACS Globalink internship to Pulkit Gera. The authors thank Bowei Chen for his early work on the project and David Ibarzabal for his help with data capture. We also thank Yohan Poirier-Ginter and Jinsong Zhang for their help.

\appendix
\section{Supplementary}

\subsection{Datasets \& Training LDR2HDR}
We train the LDR2HDR module on Laval Indoor Dataset. We augment the training set with random rotations (about the vertical axis), intensity changes (multiply the image by $2^\gamma$, with $\gamma \sim \mathrm{U}(-0.1,0.1)$) and exposure changes (make the median intensity of image $0.5+\gamma$). Here, $\mathrm{U}(a, b)$ indicates a uniform distribution in the $[a, b]$ interval. After augmentation, the resulting HDR panorama is used as target $\mathbf{t}$ for training. The input $\mathbf{x}$ is created by clipping $\mathbf{t}$ to the [0,1] interval. We further apply hue shift and unsharp mask (amount =1, $\sigma \sim \mathrm{U}(0,3)$), add small amount of per-pixel Gaussian noise ($\sigma = 0.01$), and augment the tonemapping $\mathbf{x}^{1+\gamma}$ to simulate the behaviour of a real camera. We train the network for 1500 epochs using the Adam~\cite{Kingma2015AdamAM} optimizer with a learning rate $\eta=10^{-4}$, $\beta_1=0.9$, $\beta_2=0.999$ and $\epsilon=10^{-8}$. 

Further, we finetune the model on a small dataset captured using Ricoh Theta Z to alleviate domain gap. We run 130 additional epochs using the Adam~\cite{Kingma2015AdamAM} optimizer with the same parameters as above.
 
\subsection{Training \methodname}
We use NeRF++~\cite{nerf++} as the basis of our project. An eight-layer MLP with 256 channels is used for predicting radiance and densities at the sampled points. Along each ray, we sample 64 points for training the coarse network and 128 points for training the fine network. The batch size of rays is 1024. We use integrated positional encoding to encode the inputs of the network as used in MipNeRF~\cite{mipnerf}. Similarly a single MLP is used to encode the scene. In addition, we also use spherical sampling, which weights pixels at the poles less with respect to pixels in the middle. The network is trained using the Adam optimizer~\cite{Kingma2015AdamAM} with learning rate $\eta=10^{-4}$, $\beta_1=0.9$, $\beta_2=0.999$ and $\epsilon=10^{-8}$. The resolution of the training images is $960\times480$. The network is trained for approximately 500,000 iterations, which takes around 36 hours on a NVIDIA V100 GPU.  

\subsection{Spherical Sampling}
Equirectangular images correspond to the projection of a spherical signal onto a 2D plane, where (normalized) pixel coordinates $(i,j)$ are related to azimuth $\varphi \in [-\pi, \pi]$ and elevation $\theta \in [-\nicefrac{\pi}{2}, \nicefrac{\pi}{2}]$ angles by
\begin{equation}
i = \frac{1}{2\pi} \varphi \cos\theta + \frac{1}{2} \,, \; \mathrm{and} \; j = \frac{\nicefrac{\pi}{2}-\theta}{\pi} \,.
\end{equation}
To train PanoHDR-NeRF, we sample rays in spherical coordinates instead of pixel coordinates, where $\theta \sim \mathrm{U}(-\pi, \pi)$, $\varphi = \cos^{-1}(2\beta - 1)$, and $\beta \sim \mathrm{U}(0, 1)$. \\
We compare the results between planar and spherical sampling used for training \methodname. We observe that spherical sampling performs much better and provides sharper results. 

\begin{table}
\centering

  \begin{tabular}{lSSSS}
    \toprule
    & \multicolumn{2}{c}{Planar} & \multicolumn{2}{c}{Spherical} \\
    & PSNR$\uparrow$ & SSIM$\uparrow$ & PSNR$\uparrow$  & SSIM$\uparrow$ \\
    \midrule
    Chess room & 29.635 & 0.919 & 31.389 & 0.929 \\
    Stairway & 25.381 & 0.892 & 27.381 & 0.891  \\
    Cafeteria & 22.940 & 0.845 & 24.038 & 0.842 \\
    Spotlights & 25.557 & 0.847 & 26.128 & 0.852 \\
    Dark class & 29.955 & 0.911 & 31.368 & 0.917  \\
    Small class & 29.231 & 0.906 & 30.611 & 0.911  \\
    \midrule
    Overall & 27.115 &0.886 & 28.486 &0.902 \\
    \bottomrule
  \end{tabular}
  \caption{Quantitative comparison between planar and spherical sampling (on LDR images only) averaged over all captured scenes. Spherical sampling has better results.}
  \label{tab:sample}
\end{table}

\subsection{Video}
We provide an additional video to showcase our results. We relight 3 virtual test objects made up of metal (armadillo), diffuse (bunny) and glass (sphere) with our recovered HDR images at novel viewpoints to demonstrate the dynamic range recovered. We present our results with a variety of scenes captured casually.
\newpage

\end{document}